\documentclass[final]{cvpr}

\usepackage{nopageno}
\usepackage{times}
\usepackage{epsfig}
\usepackage{graphicx}
\usepackage{amsmath}
\usepackage{amssymb}

\usepackage[export]{adjustbox}
\usepackage[nolist]{acronym}

\usepackage[utf8]{inputenc} 
\usepackage[T1]{fontenc}    
\usepackage{url}            
\usepackage{booktabs}       
\usepackage{amsfonts}       
\usepackage{nicefrac}       
\usepackage{microtype}      
\usepackage[ruled,vlined]{algorithm2e}
\usepackage{verbatim}
\usepackage{braket}
\usepackage{paralist}
\usepackage{multirow}
\usepackage{subcaption}
\usepackage{xcolor}
\usepackage{soul}
\usepackage{etoolbox}
\usepackage{caption}
\usepackage{float}
\usepackage{enumitem}
\usepackage{slashbox}
\usepackage{amsthm}

\usepackage[symbol]{footmisc}

\usepackage[normalem]{ulem}
\AtBeginEnvironment{flalign}{\setcounter{equation}{0}}
\SetKwComment{Comment}{$\triangleright$\ }{}

\newtheorem{definition}{Definition}

\usepackage[pagebackref=true,breaklinks=true,colorlinks,bookmarks=false]{hyperref}

\def\cvprPaperID{7797} 
\def\confYear{CVPR 2021}

\newcommand{\ckl}{C_\text{KL}}
\newcommand{\cfro}{C_\text{Fro}}
\newcommand{\skl}{SWBN-KL}
\newcommand{\sfro}{SWBN-Fro}
\newcommand{\red}{\textcolor{red}}

\begin{acronym}
    \acro{BN}[BN]{Batch Normalization}
    \acro{DNN}[DNN]{Deep Neural Network}
    \acro{ICS}[ICS]{Internal Covariate Shift}
    \acro{PCA}[PCA]{Principal Component Analysis}
    \acro{SGD}[SGD]{Stochastic Gradient Decent}
    \acro{SWBN}[SWBN]{Stochastic Whitening Batch Normalization}
    \acro{ZCA}[ZCA]{Zero-phase Component Analysis}
    \acro{DBN}[DBN]{Decorrelated Batch Normalization}
    \acro{ITERNORM}[IterNorm]{Iterative Normalization}
    \acro{KL}[KL]{Kullback–Leibler}
    \acro{SVD}[SVD]{Singular Value Decomposition}
\end{acronym}

\begin{document}

\title{Stochastic Whitening Batch Normalization}

\author{Shengdong Zhang$^1$ 
\and
Ehsan Nezhadarya$^{*1}$ 
\and
Homa Fashandi$^{*1}$
\and
Jiayi Liu$^{\# 2}$
\and
Darin Graham$^1$ \qquad Mohak Shah$^2$ \\
$^1$Toronto AI Lab, LG Electronics Canada \qquad
$^2$America R\&D Lab, LG Electronics USA\\
\{\tt\footnotesize{shengdong.zhang, ehsan.nezhadarya, homa.fashandi, jason.liu,  darin.graham, mohak.shah\}@lge.com}
}

\maketitle

\begin{abstract}
   \ac{BN} is a popular technique for training \acp{DNN}. \ac{BN} uses scaling and shifting to normalize activations of mini-batches to accelerate convergence and improve generalization. The recently proposed \ac{ITERNORM} method improves these properties by whitening the activations iteratively using Newton's method. However, since Newton's method initializes the whitening matrix independently at each training step, no information is shared between consecutive steps. In this work, instead of exact computation of whitening matrix at each time step, we estimate it gradually during training in an online fashion, using our proposed \ac{SWBN} algorithm. We show that while \ac{SWBN} improves the convergence rate and generalization of \acp{DNN}, its computational overhead is less than that of IterNorm. Due to the high efficiency of the proposed method, it can be easily employed in most \ac{DNN} architectures with a large number of layers. We provide comprehensive experiments and comparisons between \ac{BN}, \ac{ITERNORM},  and \ac{SWBN} layers to demonstrate the effectiveness of the proposed technique in conventional (many-shot) image classification and few-shot classification tasks.
\footnote[0]{* Equal contribution. \# Jiayi is now with Kwai Inc.}
\end{abstract}
\vspace{-10pt}
\section{Introduction}

Gradient descent-based methods are the de-facto training algorithms for \ac{DNN}, and mini-batch \ac{SGD} has become the most popular first-order optimization algorithm. In mini-batch \ac{SGD}, instead of computing the gradients for the entire training set as in batch gradient descent, or based on one training sample as in conventional \ac{SGD}, the gradients are computed based on a small random subset of the training set called mini-batch. The stochastic nature of mini-batch \ac{SGD} helps a \ac{DNN} find better local optima or even the global optima than batch gradient descent. We use \ac{SGD} to refer to mini-batch SGD in the rest of the paper. 

Due to the change in the distribution of the inputs of \ac{DNN} layers at each training step, the network experiences \ac{ICS} as defined in the seminal work of ~\cite{ioffe2015batch}. \ac{ICS} affects the input statistics of the subsequent layers, and as a result, it degrades the training efficiency. 

Eliminating the effect of \ac{ICS} can accelerate the training of \ac{DNN} by helping the gradient flow through the network, stabilizing the distributions of the activations, and enabling the use of a larger learning rate. To alleviate these effects, \ac{BN} has been proposed in~\cite{ioffe2015batch}.

Recent studies have shown that whitening (decorrelating) the activations can further reduce the training time and improve the generalization~\cite{Desjardins05Natural,Luo2017LearningDA,huang2019iterative}. However, when dealing with high-dimensional data, the requirement of eigen-decomposition~\cite{Desjardins05Natural,Luo2017LearningDA}, \acf{SVD}, or Newton's iteration~\cite{huang2019iterative} for computing whitening matrices has been the bottleneck of these methods.

\begin{figure}[t]
  \centering
\resizebox{\columnwidth}{!}{
    \includegraphics{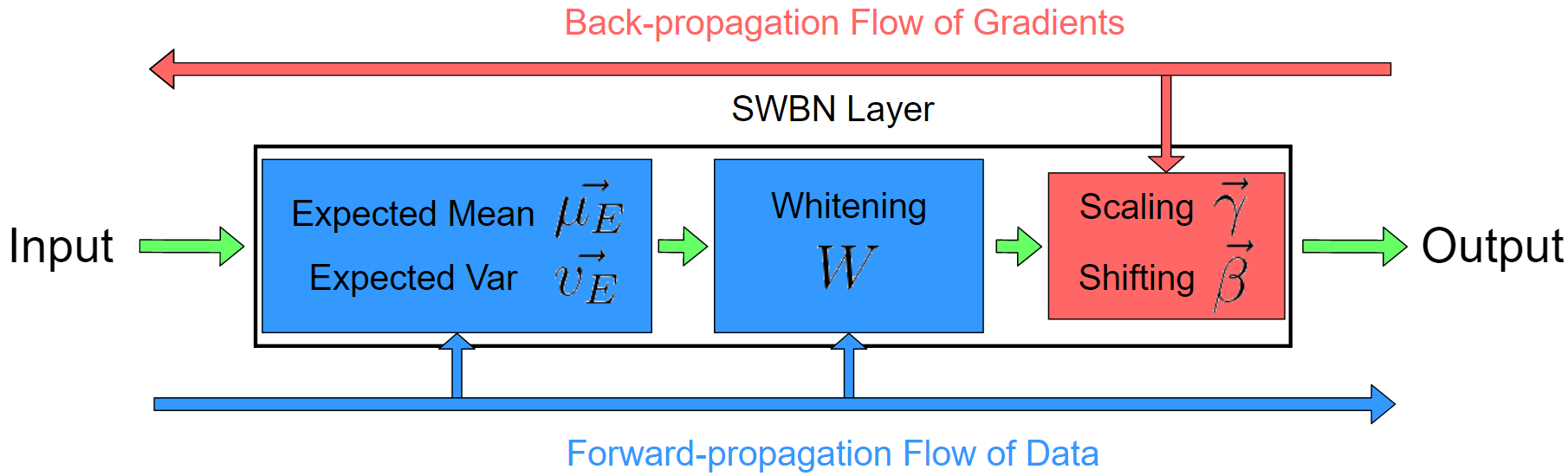}}
\setlength{\belowcaptionskip}{-10pt}
  \caption{\small The \ac{SWBN} diagram shows how whitening parameters and task parameters inside an \ac{SWBN} layer are updated in a decoupled way. Whitening parameters (in blue rectangles) are updated only in the forward phase, and are fixed in the backward phase. Task parameters (in red rectangles) are fixed in the forward phase, and are updated only in the backward phase.}
  \label{fig:swbn_diag}
\end{figure}

In this paper, we propose a method called \ac{SWBN}, which gradually learns whitening matrices separately in each layer during training. The proposed method eliminates the need for expensive matrix decomposition and inversion for whitening data as \ac{SWBN} learns the whitening matrix in an online fashion. Our formulation clearly shows the computational advantage of \ac{SWBN} to other whitening techniques applied in \acp{DNN}, such as~\cite{Desjardins05Natural}, \cite{huang2019iterative} and \cite{huang2018decorrelated}.

In Section~\ref{sec:related_work}, we review the related works on normalization and whitening techniques for \ac{DNN}. In Section~\ref{sec:SWBN}, we derive and discuss the \ac{SWBN} algorithm. In Section~\ref{sec:exp}, we present extensive experimental results to show the effectiveness of the proposed method for different network architectures and datasets. In summary, the key advantages of the proposed method are as follows:
\begin{itemize}
\item \ac{SWBN} is designed to be a drop-in replacement for \ac{BN}. It decouples the learning process of whitening matrices from the back-propagation algorithm. Thus there is no need to modify the network architecture, task loss, or the optimization algorithm to enable whitening.
\item \ac{SWBN} takes advantages of both \ac{BN} and whitening to improve convergence rate and generalization performance of \ac{DNN} models with small computational overhead.
\item There are a few whitening approaches that aim to serve as drop-in replacements for \ac{BN} layers. However, they can only replace a small number of \ac{BN} layers in a \ac{DNN} with their whitening layers due to the computational burden. In contrast, \ac{SWBN} learns whitening matrices without expensive matrix decomposition or matrix inversion operations, enabling \ac{SWBN} layers to replace a large number of \ac{BN} layers.

\end{itemize}



\section{Related Works}
\label{sec:related_work}
Activation normalization is known to be a very effective technique in speeding up the training of \acp{DNN}~\cite{Raiko2012DeepLM}. A straightforward solution is to center the input mean to zero, and use nonlinear activation functions that range from -1 to 1 \cite{LeCun1998EfficientB}. Although this method limits the activation values, activation distribution may still largely vary during the training, as caused by the change in the parameters of the previous layers known as \ac{ICS}.

To alleviate \ac{ICS}, \ac{BN} was proposed as a technique that employs the mean and variance of the mini-batch to normalize the activations~\cite{ioffe2015batch}. At training time, this technique stabilizes the distributions of the activations and allows the use of larger learning rates. \ac{BN}, however, may not work well with small mini-batches, because mean and variance estimates are less accurate. To improve the estimation accuracy for small mini-batches, other variants have been proposed, such as Batch Renormalization~\cite{Ioffe2017BatchRT}, Weight Normalization~\cite{Salimans2016WeightNA}, Layer Normalization~\cite{Ba2016LayerN}, Group Normalization~\cite{GroupNorm2018}, Online Normalization~\cite{chiley2019online}, PowerNorm~\cite{shen2020powernorm}, and Streaming Normalization \cite{Liao2016StreamingNT}, etc.

Whitening (decorrelating) the activations may further improve the training time and generalization of \ac{DNN} models. This usually happens as a result of improving the conditioning of the input covariance matrix, which leads to better conditioning of the Hessian matrix or Fisher information matrix of network parameters. One way to whiten activations is to add regularization terms to the task loss. Cogswell et al.~\cite{Cogswell2015} have proposed a regularizer, {\it DeCov}, that adds decorrelation loss to the task loss function, which encourages decorrelation between activations and non-redundant representations in \acp{DNN}. An extended version of {\it DeCov} based on group-based decorrelation loss has been proposed in \cite{Xiong2016}. In~\cite{yoshida2017spectral}, spectral norm regularization is proposed. It designs a penalty term added to the task loss function, which regularizes parameter matrices by the approximated largest singular values and their corresponding singular vectors via the Power Iteration method. Although these methods are shown to improve network generalization, they introduce extra hyper-parameter to merge the decorrelation regularizer with the task loss.

Some approaches do not introduce any extra term to the task loss. Natural neural networks~\cite{Desjardins05Natural} and generalized whitened neural networks~\cite{Luo2017LearningDA} propose two types of \ac{ZCA}-based whitening layers to improve the conditioning of the Fisher information matrix of a \ac{DNN}. To avoid the high cost of eigenvalue decomposition that is required to compute the whitening matrix, both methods amortize the cost over multiple consecutive updates, by performing the whitening step only at every certain number of iterations. Similarly, \ac{DBN}~\cite{huang2018decorrelated} directly whitens the activations by eigenvalue decomposition of the sample covariance matrix computed over each mini-batch, but its usage in a large \ac{DNN} is limited due to its high computational cost. Further, \cite{wang2019backpropagation} proposes back-propagation friendly eigen-decomposition to whiten the activations by combining the Power Iteration method and the truncated \ac{SVD}. \ac{ITERNORM}~\cite{huang2019iterative}, as an improved version of \ac{DBN}, uses Newton's method to compute the square root inverse of the covariance matrix, iteratively. Although Newton's method improves the whitening efficiency, \ac{ITERNORM} still has the following drawbacks: First, \ac{ZCA} whitening is used for each mini-batch independently. In other words, at each training step, the exact whitening matrix is computed specifically for the current mini-batch. As a result, the information of this whitening matrix does not carry over to the computation of the whitening matrix at the next step. Second, because Newton's method requires multiple iterations of matrix multiplication, applying \ac{ITERNORM} to all the layers of a very deep neural network is computationally inefficient.

\section{Stochastic Whitening Batch Normalization}
\label{sec:SWBN}

In this section, we first review the whitening techniques and explain the issues that arise when these techniques are employed directly in the training of \ac{DNN} models. Later, we introduce our stochastic whitening batch normalization algorithm and provide explanations and complexity analysis of how and why \ac{SWBN} works.

\subsection{Whitening Transformation}
\label{sec:white_transform}
A random vector $\vec{z} \in \mathbb{R}^{d}$, with zero mean, is said to be white if the expectation of the covariance of $\vec{z}$ satisfies $E[\vec{z}\vec{z}^{T}] = I_d$, where $I_d$ is an identity matrix. Therefore, elements of $\vec{z}$ have unit variance and are mutually uncorrelated. Whitening is a process that transforms a zero-mean random vector $\vec{x} \in \mathbb{R}^d$ into $\vec{z}$ by a linear transformation. Existing methods for data whitening are to search for a transformation matrix $W\in\mathbb{R}^{d\times d}$, such that $\vec{z}=W\vec{x}$~\cite{kessy2018optimal}. \ac{PCA} whitening and \ac{ZCA} whitening algorithms are two commonly used methods. Both of these algorithms require the covariance matrix $\Sigma_{\vec{x}} = E[\vec{x}\vec{x}^T]$, and its decomposition via eigenvalue decomposition, or Cholesky decomposition on its inverse matrix $\Sigma_{\vec{x}}^{-1}$.

The eigenvalue decomposition decomposes the covariance matrix as $\Sigma_{\vec{x}} = UDU^T$, where $U$ is an orthogonal matrix and $D$ is a diagonal matrix with eigenvalues of the covariance matrix on its diagonal. In the case of Cholesky decomposition, we have $\Sigma_{\vec{x}}^{-1}=L L^T$, where $L$ is a lower triangular matrix with positive diagonal values. In practice, the sample covariance matrix $\hat{\Sigma}_{\vec{x}}$ is used. In this work, we only consider \ac{PCA} and \ac{ZCA} whitening algorithms. In \ac{PCA} whitening, the transformation matrix $W$ is of the form $W = D^{-1/2}U^T$, and in \ac{ZCA} whitening $W$ is of the form $W = UD^{-1/2}U^T$. It is worth mentioning that left-multiplying any orthogonal matrix to the \ac{PCA} whitening matrix $W$ forms a new whitening matrix \cite{kessy2018optimal}. The \ac{ZCA} whitening matrix is the only whitening matrix that is symmetric. 

More details of these whitening algorithms are included in~\cite{bell1997independent}, \cite{eldar2003mmse}, \cite{jolliffe2011principal}, and~\cite{kessy2018optimal}. 

\subsection{Introduction to \ac{SWBN}}

The computational cost of whitening matrices usually becomes the bottleneck when applying any of the above-mentioned whitening algorithms to train a \ac{DNN}, especially for networks with millions or billions of parameters. One important question to answer is whether the complete whitening process is necessary at each training step. Because of \ac{ICS}, a whitening matrix computed at one step could be very different from the one computed at the next step, making the full computation of the whitening matrix at the previous step a waste. Therefore, it will be ideal if the whitening algorithm can reduce the computational cost via gradually whitening the data over training iterations.

In this work, we introduce \ac{SWBN}, a stochastic algorithm that gradually learns whitening matrices and whitens activations simultaneously. \ac{SWBN} whitens the activations by stochastically minimizing a whitening loss with respect to an internal matrix. A whitening loss is a function of the covariance matrix. The internal matrix keeps track of the changes of the input distribution through the loss minimization, and eventually becomes a whitening matrix. However, unlike DeCov, spectral norm, or any other methods involving modification of the loss functions, in \ac{SWBN}, the whitening loss is decoupled from the task loss. Decoupling them not only reduces the chance of divergence at training time but also speeds up convergence. Also, as shown in \cite{huang2019iterative}, although fully whitening the activations helps accelerate convergence, partial whitening on each mini-batch may yield better generalization due to the noise introduced from partial whitening. In addition, different from \ac{DBN} or \ac{ITERNORM} that completely whiten activations at each step, an SWBN layer uses its internal matrix to “slightly” whiten the activations with respect to a predefined whitening loss before they are fed into the next layer. As training continues, the matrix gets closer to the final whitening matrix, and the output of an \ac{SWBN} layer becomes whiter. We discuss two whitening criteria in the next subsection.

\subsection{Whitening Criteria}

We define the whitening criterion as the whitening loss function, which is a measure of distance between a covariance matrix and the identity matrix.

\begin{definition}
A whitening criterion for a positive semi-definite matrix $\Sigma$ is a function $C: \mathbb{S}^n_+ \mapsto \mathbb{R}^+\cup \{0\} $ that maps $\Sigma$ to a non-negative real number which quantifies the dissimilarity between $\Sigma$ and the identity matrix $I$. $\mathbb{S}^n_+$ represents a set of positive semi-definite matrices of size $n\times n$. 
\end{definition}
With this definition, we can define a whitening matrix under a criterion $C$ for a random vector $\vec{x}$.
\begin{definition}
Let $\vec{x} \in \mathbb{R}^d$ be a zero-mean $d$-dimensional random vector and $\Sigma_{\vec{x}}=E[\vec{x}\vec{x}^T]$. A matrix $W^*$ is called a whitening matrix under a criterion C, or C-whitening matrix, of $\vec{x}$, $W$ satisfies:
{\small \[ W^* = \arg\min_{W} \,\,C(\Sigma_{\vec{y}}) = \arg\min_{W} C(W\Sigma_{\vec{x}}W^T) \]}
\end{definition}

\vspace{-5pt}
\noindent where {\small $\vec{y}=W\vec{x}$}, and {\small $\Sigma_{\vec{y}} = E[\vec{y}\vec{y}^T]=WE[\vec{x}\vec{x}^T]W^T$}.

In this work, we consider the following two whitening criteria derived by \ac{KL} divergence and Frobenius norm:
{\footnotesize
\begin{flalign}
\ckl(W\Sigma_{\vec{x}}W^T) &= \frac{1}{2}(tr(W\Sigma_{\vec{x}}W^T)-\ln\det(W\Sigma_{\vec{x}}W^T)-d)\label{KL_criterion} \\
\cfro(W\Sigma_{\vec{x}}W^T) &= \frac{1}{2}||I - W\Sigma_{\vec{x}}W^T ||_\text{Fro}\label{F_criterion}
\end{flalign}
}
It is obvious that both criteria reach their minimum values of $0$ if {\small $W\Sigma_{\vec{x}}W^T=I_d$}. The first criterion is derived from \ac{KL} divergence based on the assumption of having two zero-mean Gaussian distributions with covariance matrices equal to $\Sigma_{\vec{y}}$ and $I_d$. The second criterion directly computes the Frobenius norm of the difference between the identity matrix $I_d$ and the sample covariance matrix. Unlike $C_{KL}$, $C_{Fro}$ has no assumptions on the probability distribution. These two criteria are the core of the proposed \ac{SWBN} algorithm. More details can be found in Appendix B.

\subsection{Update Rules for \ac{SWBN} Layer}

Assume $\vec{x}$ is an input vector to a hidden layer of a \ac{DNN} model. We find the $C$-whitening matrix $W$ of each layer by minimizing the whitening criterion using \ac{SGD}. We update $W$ by $W \gets W - \alpha\Delta W$, where $\alpha$ is the step size and $\Delta W$ is the update matrix. The update rules with respect to the criteria in Eq.~($1$) and Eq.~($2$) mentioned above are:
\begin{equation}
\Delta W = \left\{\begin{matrix}
(W\hat{\Sigma}_{\vec{x}}W^T - I_d)W, \quad \hfill{\text{for $\ckl$}}, \\  
\frac{(W\hat{\Sigma}_{\vec{x}}W^T - I_d)W\hat{\Sigma}_{\vec{x}}}{||I_d - W\hat{\Sigma}_{\vec{x}}W^T ||_{Fro}}, \quad \hfill{\text{for $\cfro$}},
\end{matrix}\right.
\label{eq:update}
\end{equation}
where $\hat{\Sigma}_{\vec{x}}$ is the sample covariance matrix.

Cardoso et al.~\cite{cardoso1996equivariant} shows that optimizing $W$ to be a minimizer of $\ckl$ by its update rule in Eq.~(\ref{eq:update}) results in a whitening matrix. To our knowledge, this is the first time that the update rule of $C_{KL}$ is applied in a mini-batch SGD setting for training \ac{DNN}s. The proposed update rule of $C_{Fro}$ is not only less sensitive to the whitening step size and the batch size, but also shows better performance on few-shot classification, based on the experimental results in Section~\ref{sec:exp}. Unlike the update rule of $\cfro$, the update rule of $\ckl$ is derived by relative gradients. The derivation of these update rules and the detailed discussion are given in Appendix B.

\SetKwInOut{Input}{Input}
\SetKwInOut{Output}{Output}

\subsection{\ac{SWBN} Algorithm}

\begin{algorithm}[t]
\caption{Forward Propagation of SWBN Layers at Training Phase}
\label{alg:train}
\DontPrintSemicolon
\Input{\small{ Input batch} $X \in\mathbb{R}^{d\times n} = [\vec{x}_1, \cdots, \vec{x}_n]$}
\Output{\small{ Processed data} $\hat{X}\in \mathbb{R}^{d\times n}$} 
{\bf Initialization:} \\
\textit{Whitening parameters}: $W= I_d$; \\
\textit{Task parameters:} $\vec{\gamma}$, $\vec{\beta} \in \mathbb{R}^d$; \\
\textit{Hyperparameters:} tolerance $\epsilon=10^{-8}$, moving average momentum $\eta=0.95$, step size $\alpha$; \\ 
\textit{Criterion}: $C=\ckl$ or $\cfro$; \\
\textit{Expected mean and variance}: $\vec{\mu}_{E}= \vec{0}$, $\vec{v}_E= \vec{1}$.\\
\vspace{2pt}
1. \small{Calculate batch mean:} $\vec{\mu} \gets \frac{1}{n}\Sigma_{i=1}^n \vec{x}_i$

2. \small{Calculate batch variance:} $\vec{v} \gets \frac{1}{n-1}\Sigma_{i=1}^n (\vec{x}_i-\vec{\mu})\odot(\vec{x}_i-\vec{\mu})$\footnotemark{}

3. \small{Update the expected mean:} $\vec{\mu}_E \gets \eta\vec{\mu}_E + (1-\eta)\vec{\mu}$

4. \small{Update the expected variance:} $\vec{v}_E \gets \eta\vec{v}_E + (1-\eta)\vec{v}$

5. \small{Standardize data:} $X^{S} \gets \Lambda^{-\frac{1}{2}}(X - \vec{\mu} \vec{1}_n^T)$, where $\Lambda \gets diag(\vec{v}) + \epsilon I_d$\footnotemark{}

6. \small{Calculate sample covariance matrix:} $\hat{\Sigma}_B \gets\frac{1}{n}X^{S}(X^{S})^T$

7. \small{Compute and update matrix from Eq.~(\ref{eq:update}):} $W \gets W - \alpha \Delta W$

8. \small{Enforce symmetry constraint:} $W \gets 0.5(W + W^T)$

9. \small{Multiply $W$ to standardized data:} $X^W \gets WX^{S}$

10. \small{Compute output:} $\hat{X} \gets X^W\odot (\vec{\gamma} \vec{1}^T_n) +  (\vec{\beta} \vec{1}^T_n)$
\end{algorithm}

\begin{algorithm}
\caption{Back-propagation of \ac{SWBN} Layers at Training Phase}\label{alg:back_prop}
\DontPrintSemicolon
\SetAlgoLined
\Input{ Gradients of task loss $L$ w.r.t output $\frac{\partial L}{\partial \hat{X}} \in \mathbb{R}^{d\times n}$}

Intermediate data from Algorithm 1, $X^{S}$, $X^W$, $\vec{\mu}$, $\vec{v}$ 

Whitening matrix from Algorithm 1, $W$

\Output{ Gradients w.r.t. $X$, $\vec{\gamma}$, $\vec{\beta}$ i.e. $\frac{\partial L}{\partial X}\in \mathbb{R}^{d\times n}$, $\frac{\partial L}{\partial \vec{\gamma}}\in \mathbb{R}^{d}$, $\frac{\partial L}{\partial \vec{\beta}}\in \mathbb{R}^{d}$}

{\bf Initialization}: $\frac{\partial L}{\partial X}\gets \mathbf{0}_{d\times n}$, $\frac{\partial L}{\partial \vec{\gamma}}\gets \vec{0}_d$, $\frac{\partial L}{\partial \vec{\beta}}\gets \vec{0}_d$

\For{$k = 1, \dots, d$}{
 $[\frac{\partial L}{\partial \vec{\gamma}}]_k \gets \Sigma_{j=1}^n\frac{\partial L}{\partial \hat{X}_{kj}}X^W_{kj}$ 
 \;
 $[\frac{\partial L}{\partial \vec{\beta}}]_k \gets \Sigma_{j=1}^n\frac{\partial L}{\partial \hat{X}_{kj}}$ 
\;
\For{$l = 1, \dots, n$}{
  $\frac{\partial X^{S}_{kl}}{\partial v_k} \gets -\frac{1}{2}(X^S_{kl}-\mu_k)(v_k + \epsilon)^{-\frac{3}{2}}$\; 
  $\frac{\partial X^{S}_{kl}}{\partial \mu_k} \gets -\frac{1}{\sqrt{v_k+\epsilon}}$\;  
  $\frac{\partial v_k}{\partial X_{kl}} \gets \frac{2}{n-1}(X_{kl}-\mu_k)$ \;
  $\frac{\partial \mu_k}{\partial X_{kl}} \gets \frac{1}{n}$ \;
  $\frac{\partial X^{S}_{kl}}{\partial X_{kl}} \gets \frac{1}{\sqrt{v_k+\epsilon}} + \frac{\partial X^{S}_{kl}}{\partial v_k}\frac{\partial v_k}{\partial X_{kl}}+ \frac{\partial X^{S}_{kl}}{\partial \mu_k}\frac{\partial \mu_k}{\partial X_{kl}}$ \;
  $[\frac{\partial L}{\partial X}]_{kl} \gets \frac{\partial X^S_{kl}}{\partial X_{kl}} \Sigma_{i=1}^d  \gamma_i W_{ik}\frac{\partial L}{\partial \hat{X}_{il}}$
 }
}
\end{algorithm}

\addtocounter{footnote}{-2}
\stepcounter{footnote}\footnotetext{$\odot$ stands for Hadamard product.}
\stepcounter{footnote}\footnotetext{$diag(\cdot)$ creates a diagonal matrix whose diagonal is the given vector.}

Forward propagation steps for \ac{SWBN} layer in training and prediction phases are shown in Algorithms~\ref{alg:train} and~\ref{alg:pred}, respectively. We define two sets of parameters in the algorithm: {\it whitening} parameters and {\it task} parameters. Figure~\ref{fig:swbn_diag} illustrates how these two sets of parameters are updated in the training and prediction phases. 

For a \ac{DNN} model with a convolutional layer, the input to an \ac{SWBN} layer is a tensor $T \in \mathbb{R}^{d\times h \times w \times n}$, where $d$, $h$, $w$ and $n$ stand for the number of feature channels, height, width and batch size, respectively. Note that in the test phase, $n$ is equal to $1$. To apply any of the above algorithms to $T$, we just need to reshape it into a matrix $X \in \mathbb{R}^{d\times(hwn)}$ before feeding it to the layer, and reshape the output $\hat{X}$ back to the original shape.


Steps $1$ to $5$ of Algorithm~\ref{alg:train} standardize each element of the input $\vec{x}$ to have zero mean and unit variance. Standardization can stabilize and improve the training convergence rate in a way similar to \ac{BN}. More importantly, it avoids potential numerical issues on covariance matrix estimation. Since the sample covariance matrix $\hat{\Sigma}$ is used to compute the update $\Delta W$, it has a direct influence on the convergence of $W$ as well as the training of the whole \ac{DNN}. If we estimate $\hat{\Sigma}$ simply by the centered data $X-\vec{\mu}\vec{1}^T$, due to the stochastic nature of training a \ac{DNN}, the numerical range of entries of $\hat{\Sigma}$ could have a large variation, especially at the early stages of training. This can make the learning of $W$ unstable, or may cause the training to diverge. \ac{ITERNORM}~\cite{huang2019iterative} avoids this problem by normalizing the sample covariance matrix by its trace. In \ac{SWBN}, standardizing $\vec{x}$ solves the problem, because the resulting sample covariance matrix $\hat{\Sigma}_B$ of the standardized data becomes the sample correlation matrix, whose entries are in the range $[-1, 1]$. In step $7$ of Algorithm~\ref{alg:train}, we use the update rule of Eq.~\ref{eq:update} to optimize $W$. For stable convergence, the step size $\alpha$ needs to be a small positive number, e.g. $10^{-5}$. Note that in the beginning of training, $W$ dose not fully whiten the data. As training continues, the updates to $W$ make it a better whitening matrix with respect to the chosen criterion $C$. As discussed in Section~\ref{sec:white_transform}, to render the \ac{ZCA} whitening matrix symmetric, we also enforce the symmetry constraint on $W$ in step $8$. In step $10$, following the same procedure as \ac{BN} to scale and shift the standardized input of each layer, we apply an affine transformation $\Gamma \in \mathbf{R}^{d\times d}$ to the whitened activations and shift it by a vector $\vec{\beta}$. To keep the number of parameters and thus the computational complexity low, we employ a diagonal matrix for $\Gamma$, as in~\cite{ioffe2015batch}, which is the same as applying the scaling factor $\gamma_i$ to each channel. 

During the back-propagating step, we only need to compute the gradients for the preceding layers $\frac{\partial L}{\partial X}$, and the gradients for the scaling and shifting parameters $\frac{\partial L}{\partial \vec{\gamma}}$ and $\frac{\partial L}{\partial \vec{\beta}}$, as described in Algorithm~\ref{alg:back_prop}. We do not compute the gradients of the whitening matrix $W$. The detailed derivation of the gradients in Algorithm~\ref{alg:back_prop} are given in Appendix C.


In the prediction phase, as described in Algorithm~\ref{alg:pred}, we standardize the input to the layer by the fixed expected mean $\vec{\mu}_E$, and the variance $\vec{v}_E$. Then we whiten the standardized data by the whitening matrix $W$ obtained from Algorithm~\ref{alg:train}. The output vector is then computed by scaling and shifting the whitened standardized input.

\begin{algorithm}[tpb]
\caption{Forward Propagation of SWBN Layers at Prediction Phase}\label{alg:pred}
\DontPrintSemicolon
\SetAlgoLined
\Input{ Input feature vector to the layer $\vec{x} \in \mathbb{R}^{d}$ }
\Output{ Output feature vector $\vec{x}' \in \mathbb{R}^{d}$}

1. $\vec{x}^{S} \gets \Lambda_E^{-\frac{1}{2}}(\vec{x} - \vec{\mu}_E)$, where $\Lambda_E \gets diag(\vec{v}_E) + \epsilon I_d$

2. $\vec{x}' \gets (W\vec{x}^{S})\odot \vec{\gamma} + \vec{\beta}$
\end{algorithm}

\subsection{Computational Complexity}
\label{sec:complexity}
We choose to compare \ac{SWBN} only with \ac{ITERNORM}, because \ac{DBN} adopts eigenvalue decomposition to compute the whitening matrix and thus has much higher computational cost. We consider the total number of multiplications required by matrix multiplications in these algorithms, as they dominate computation.

Let's assume the input data matrix is of size $\mathbf{R}^{d\times n}$, where $d$ is the number of feature channels, and $n$ is the number of data samples. At training time, the \ac{ITERNORM} algorithm has three steps that depend on matrix multiplications: 1) calculation of the sample covariance matrix, 2) Newton iterations for the whitening matrix, 3) and whitening the input data. Steps $1$ and $3$ require $2d^2n$ multiplications. The update formula of Newton iteration for the whitening matrix is given by $W_k=\frac{1}{2}(3W_{k-1} - W^3_{k-1}\hat{\Sigma}_N)$~\cite{huang2019iterative}, where $W_k\in \mathbb{R}^{d\times d}$ is the whitening matrix at the $k$th iteration, and $\hat{\Sigma}_N\in \mathbb{R}^{d\times d}$ is the sample covariance matrix normalized by its trace. The number of matrix multiplications for $T$ iterations is $3Td^3$. Thus, \ac{ITERNORM} requires $2d^2n+3Td^3$ multiplications in total.

In \ac{SWBN}, the majority of the computation during training time comes from steps $6$, $7$, and $8$ in Algorithm~\ref{alg:train}. Similar to \ac{ITERNORM}, the cost of steps $6$ and $8$ stems from the computation of the sample covariance matrix and whitening the input data. It is trivial to show that step $7$ in Algorithm~\ref{alg:train} requires $3$ matrix multiplications for both $C_{KL}$ and $C_{Fro}$, which result in the total number of multiplications $2d^2n+3d^3$. 
\ac{ITERNORM} requires $T=5$ to give stable performance, resulting in $2d^2n+15d^3$ multiplications, while SWBN-KL and SWBN-Fro need $2d^2n+3d^3$ multiplications. SWBN's constant of the leading term $d^3$ is five times smaller than that of \ac{ITERNORM}. As a result, SWBN is computationally more efficient. At inference time, similar to \ac{BN} and \ac{ITERNORM}, the \ac{SWBN} layer can be merged into its adjacent fully-connected layers or convolutional layers. Therefore, the \ac{SWBN} algorithm adds no extra computational overhead at inference time.

In addition, \ac{SWBN} is more memory efficient than \ac{ITERNORM}. As indicated in~\cite{huang2019iterative}, in the forward phase, \ac{ITERNORM} needs to store all intermediate whitening matrices $W_k$'s from the Newton iterations, as they are required to compute gradients in the backward propagation phase. \ac{SWBN} only needs to store one $W$, as this matrix is static in the backward propagation phase. In other words, for \ac{ITERNORM}, if $T=5$, then $5d^2$ memory space is needed, whereas \ac{SWBN} only takes $d^2$.

\section{Experiments}
\label{sec:exp}

In this section, we show the effectiveness of SWBN in terms of convergence speed and generalization through ablation studies and experiments on benchmark datasets for the classification task. In section~\ref{sub:ablation}, We demonstrate the effect of whitening step size $\alpha$ and the batch size on each model loss and convergence rate, both at training and test phases. Also, we conduct experiments to show how effectively the proposed \skl~and \sfro~layers can whiten the features maps. The computational complexity comparison between \ac{SWBN} and \ac{ITERNORM} is given in Section~\ref{sec:time_compare}. In Section~\ref{exp:classification}, we show that by replacing \ac{BN} layers with \ac{SWBN} layers, \ac{DNN} models achieve better generalization performance and training efficiency on benchmark classification datasets CIFAR-$10$, CIFAR-$100$~\cite{CIFAR10} and ILSVRC-$2012$~\cite{imagenet_cvpr09}, as well as few-shot classification benchmark datasets CIFAR-FS~\cite{bertinetto2018meta} and mini-Imagenet~\cite{vinyals2016matching}.

\subsection{Ablation Studies}
\label{sub:ablation}

We conduct all the experiments for ablation studies in a controlled configuration. For each experiment on a dataset, we first implement a model with \ac{BN}~layers. Next, we make exact copies of this model and replace their \ac{BN}~layers with~\skl,  \sfro~layers, and \ac{ITERNORM} layers, respectively. All the scaling and shifting parameters $\vec{\gamma}$ and $\vec{\beta}$ in \ac{SWBN} layers are initialized to $1$'s and $0$'s, respectively. This insures that all the models in an experiment have identical model parameters before training. We use \ac{SGD} with the learning rate of $0.1$ with momentum of $0.9$ as the optimization algorithm. We set the batch size to $128$ for all the experiments, unless stated otherwise. To remove any possible factor that may affect network performance other than these normalization layers, we do not use any regularization techniques, such as weight decay  or dropout~\cite{srivastava2014dropout}. CIFAR-$10$~\cite{CIFAR10} is used for the experiments, which has $60$K, $32\times 32$ pixels color images, $50$K in the training set and $10$K in the test set. The task is to classify images into $10$ categories.


\subsubsection{Effect of Whitening Step Size}

\begin{figure}
  \centering
    \includegraphics[width=0.33\linewidth]{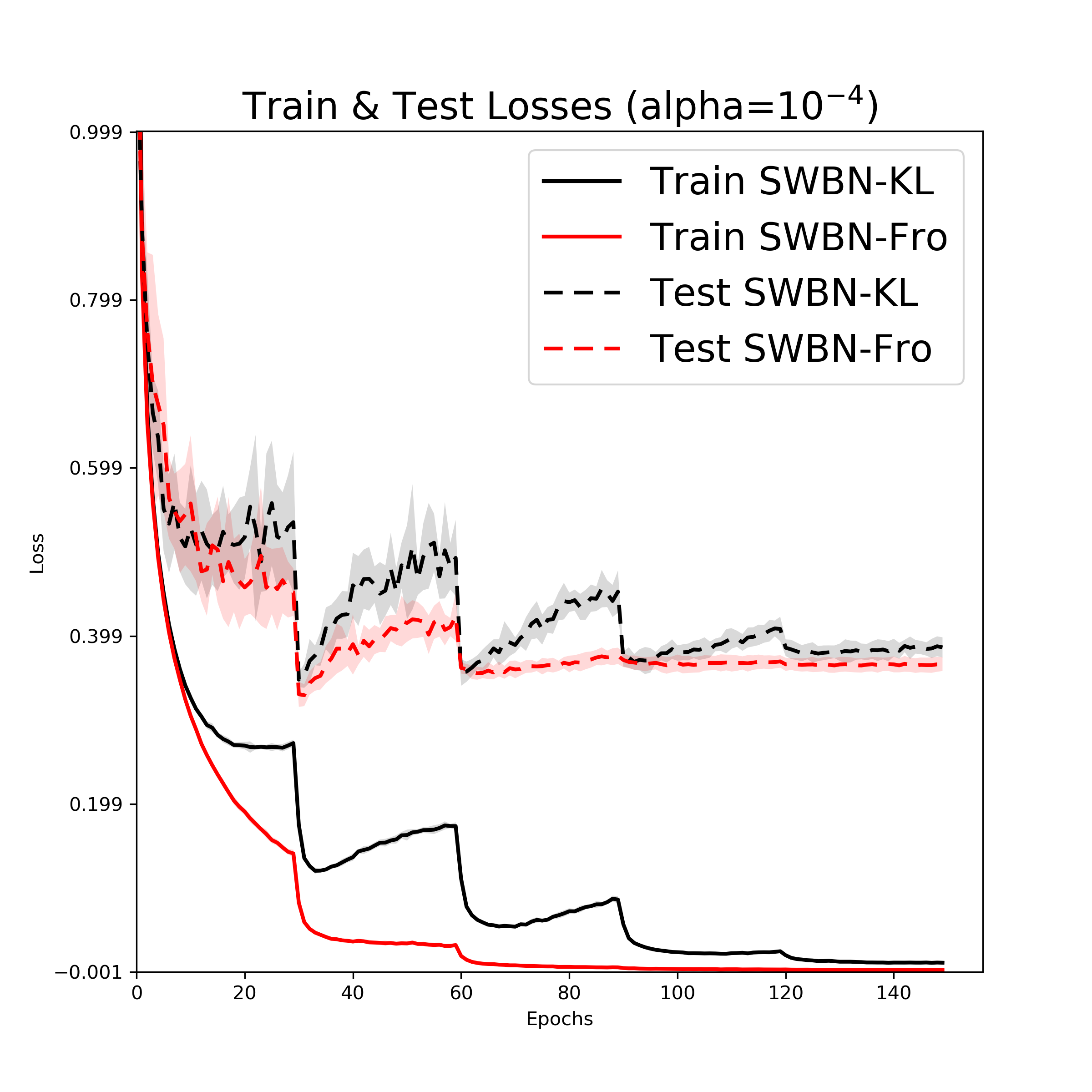}
    \hspace{-6pt}
    \includegraphics[width=0.33\linewidth]{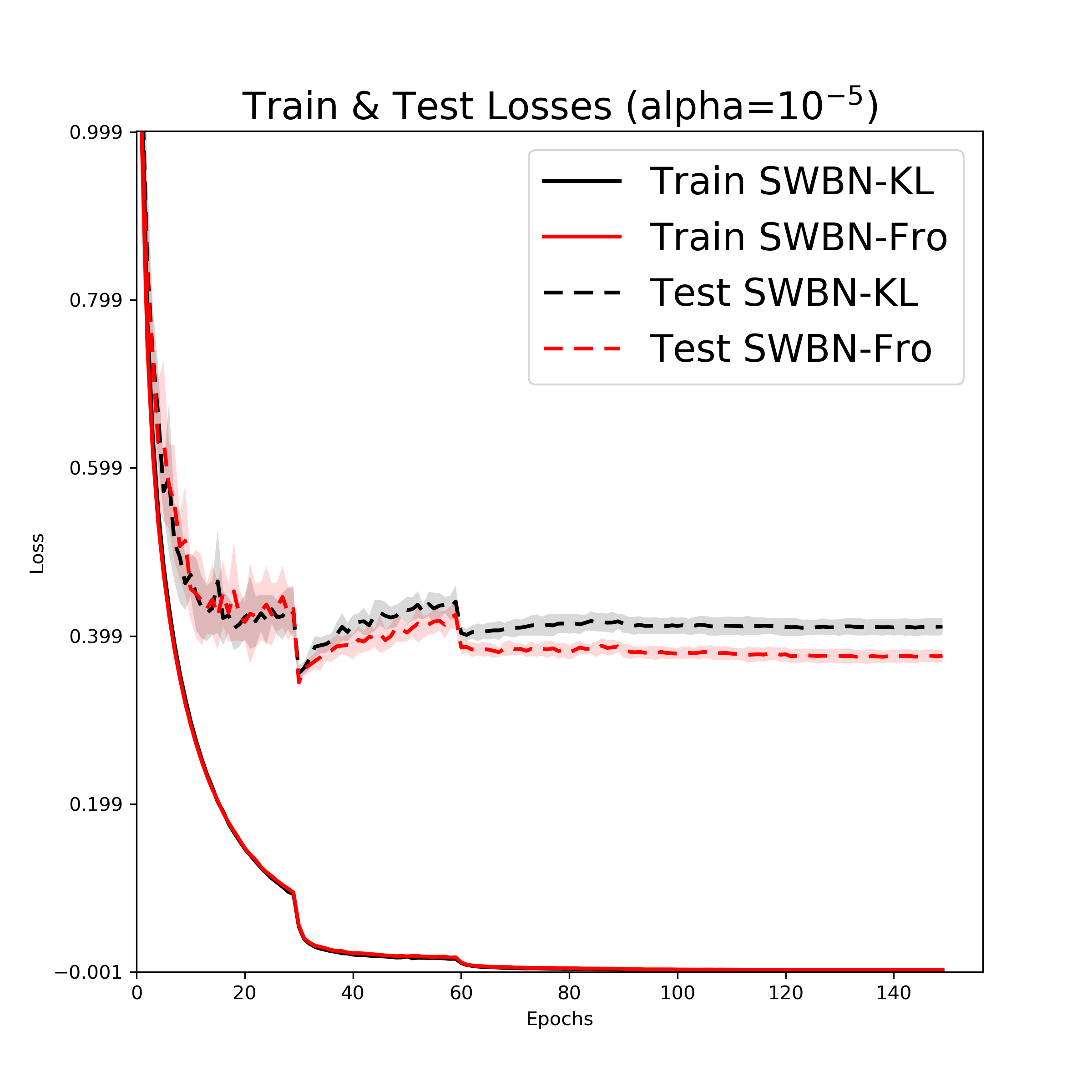}
    \hspace{-6pt}
    \includegraphics[width=0.33\linewidth]{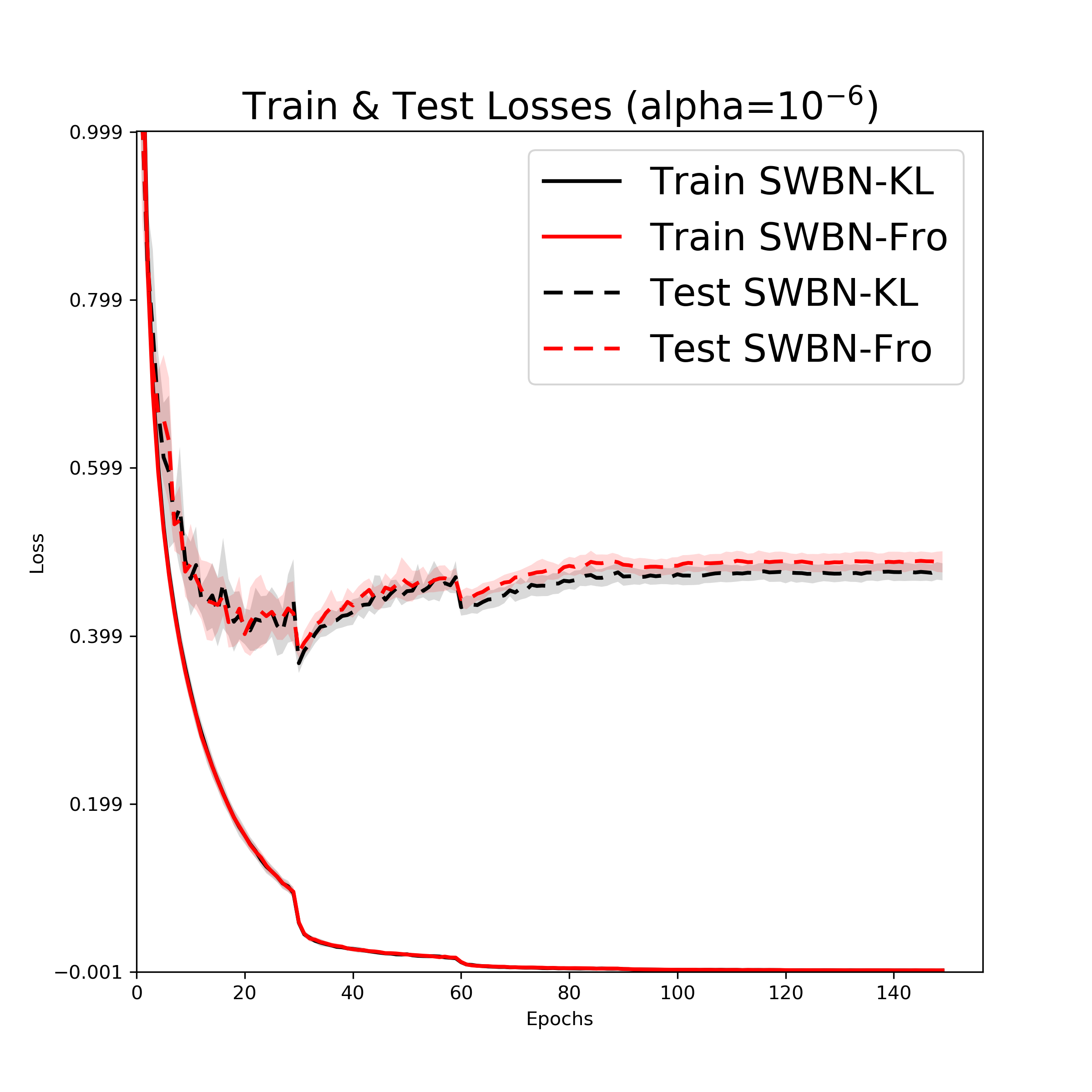}
    \setlength{\belowcaptionskip}{-10pt}
  \caption{\small Effect of whitening step sizes on training and test loss. The plots show the mean curves of $10$ runs with $\pm1$ standard deviation. Best viewed in color.}
  \label{fig:alpha_comparison}
\end{figure}

The step size $\alpha$ in Algorithm~\ref{alg:train} is an important hyper-parameter which controls the convergence speed of a whitening matrix. To investigate how this number affects the training of a model, we use a common VGG model architecture given in Appendix A, and train it on CIFAR-$10$ dataset. Each experiment is run for $100$ epochs. The learning rate is divided by $2$ at every $30$ epochs. The loss curves for $\alpha \in \{10^{-4}, 10^{-5}, 10^{-6}\}$ are depicted in Figure~\ref{fig:alpha_comparison}. For $\alpha=10^{-4}$, the convergence behavior is not as stable as that of $\alpha=10^{-5}$ and $10^{-6}$. In comparison with \skl, \sfro~shows slightly better stability. We conjecture that the Frobenius norm denominator $||I_d - W\hat{\Sigma}_{\vec{x}}W^T ||_{Fro}$ normalizes the gradients.

When $\alpha = 10^{-6}$, although the convergence is more stable than $10^{-4}$, it yields larger test loss than $\alpha = 10^{-5}$, and the generalization improvement of \ac{SWBN} seems negligible. In comparison with \skl,~\sfro~yields lower test loss. The experiment results show that $\alpha = 10^{-5}$ gives a better trade-off between convergence rate and stability.

\subsubsection{Effect of Batch Size}

\label{sec:INvsSWBN}
\begin{figure}
  \centering
    \includegraphics[width=0.33\linewidth]{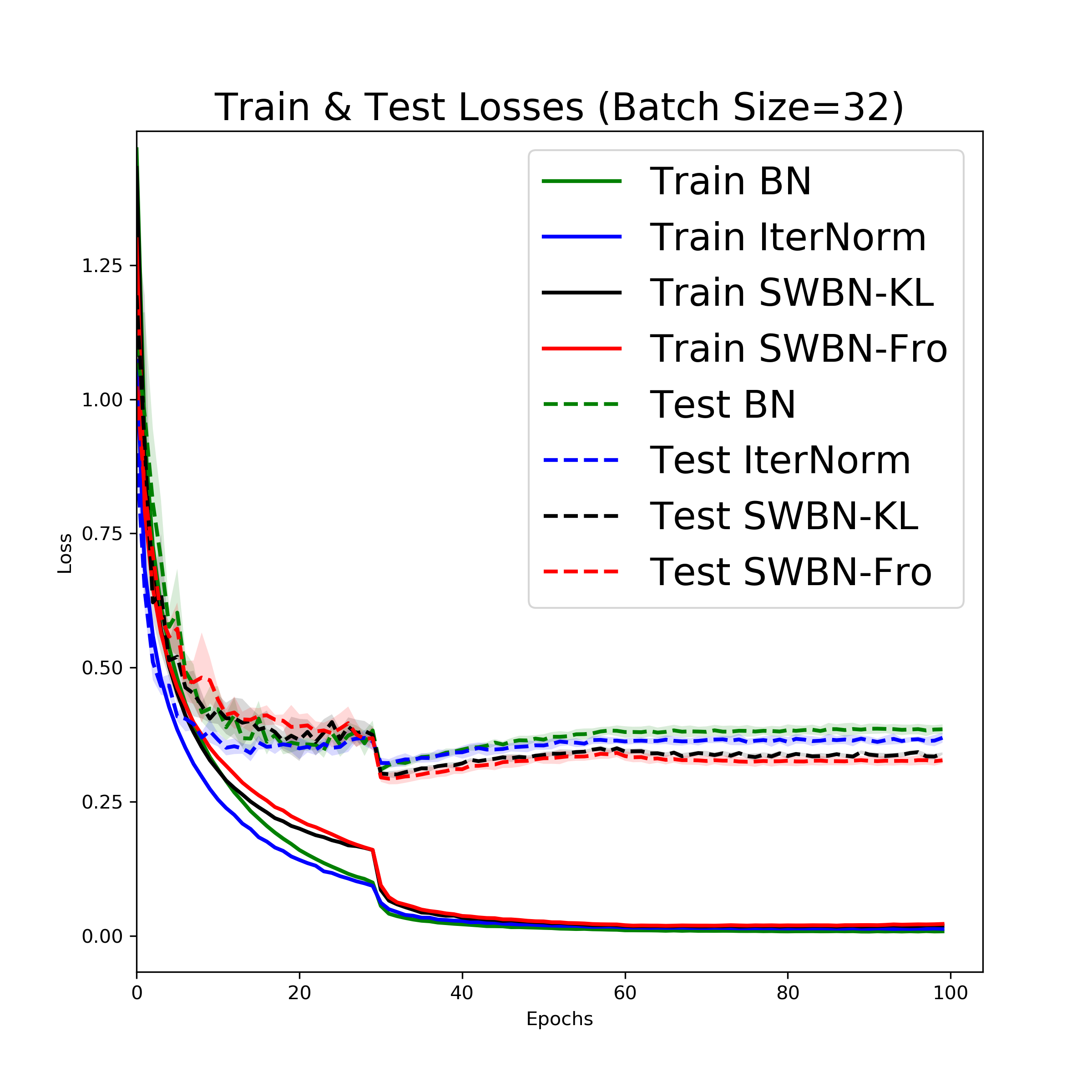}
    \hspace{-6pt}
    \includegraphics[width=0.33\linewidth]{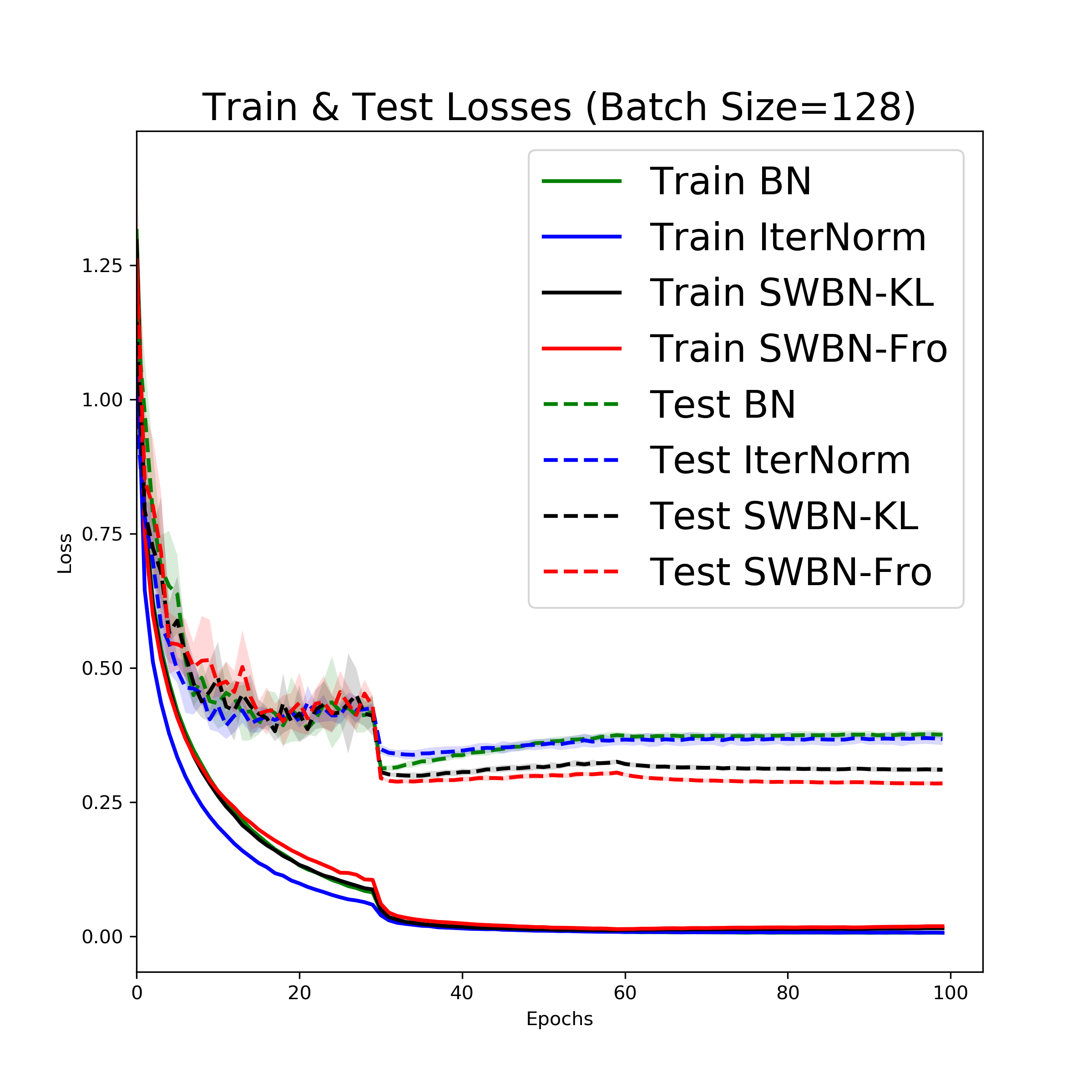}
    \hspace{-6pt}
    \includegraphics[width=0.33\linewidth]{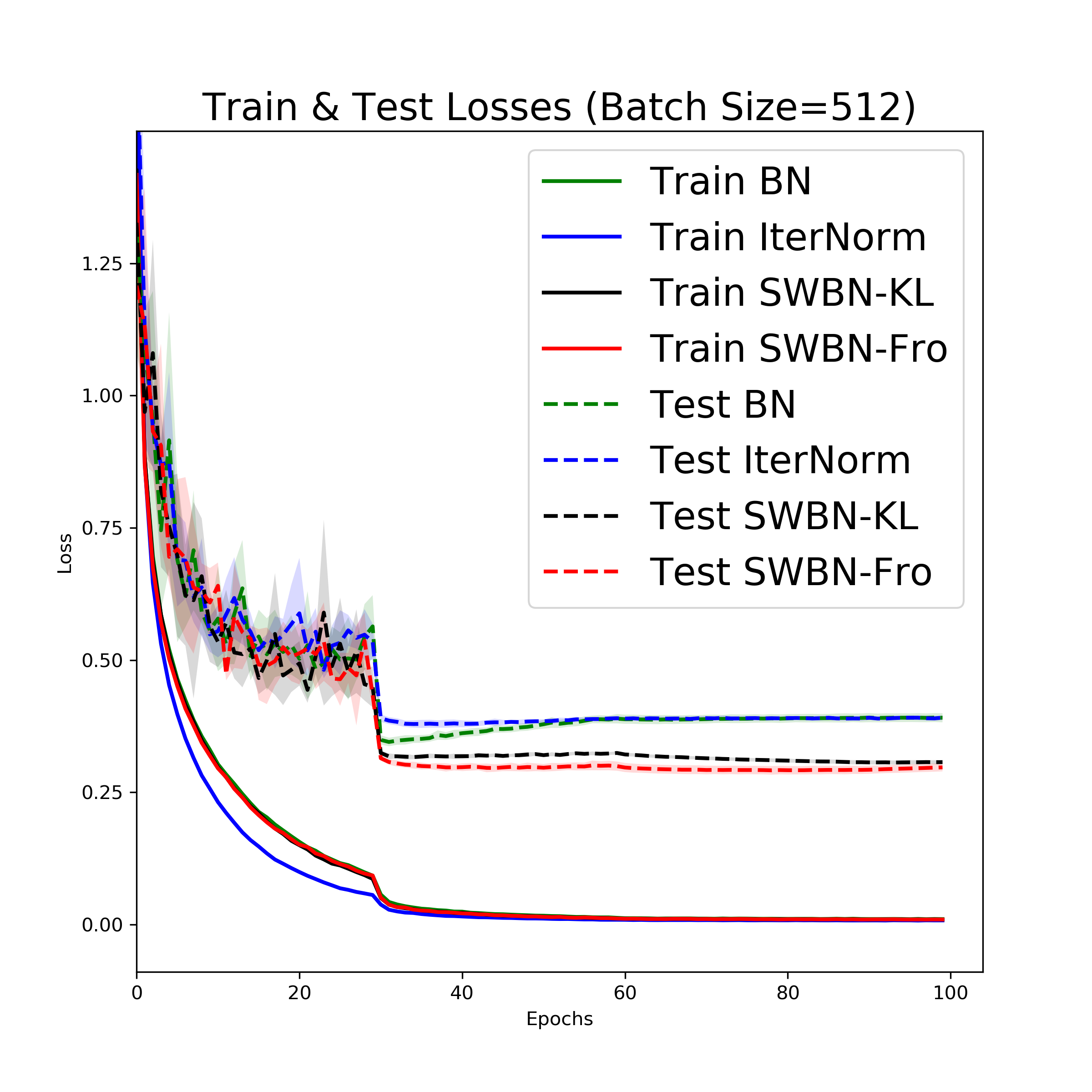}
    \setlength{\belowcaptionskip}{-10pt}
  \caption{\small Effect of different batch sizes. The plots show the mean curves of 5 runs with $\pm1$ standard deviation. Best viewed in color.}
  \label{fig:bs_comparison}
\end{figure}

We also show the effect of different batch sizes on the performance of SWBN. We also include the results of \ac{ITERNORM} for comparison. We follow the same configuration, except each experiment is run for $150$ epochs, and the learning rate is divided by $10$ at every $30$ epochs. The loss curves with $1$ standard deviation error bars for batch sizes $32$, $128$ and $512$ are shown in Figure~\ref{fig:bs_comparison}. As seen, the lowest test losses for different batch sizes are achieved by \sfro. The models with SWBN layers outperform those with \ac{BN} and \ac{ITERNORM} layers in terms of test loss. 

\subsubsection{Whitening Effect of SWBN}
\label{sec:EoW}

\begin{figure}[t]
  \begin{subfigure}{\linewidth}
  \centering
  \rotatebox{90}{\qquad Epoch $0$}
    \includegraphics[width=0.3\linewidth]{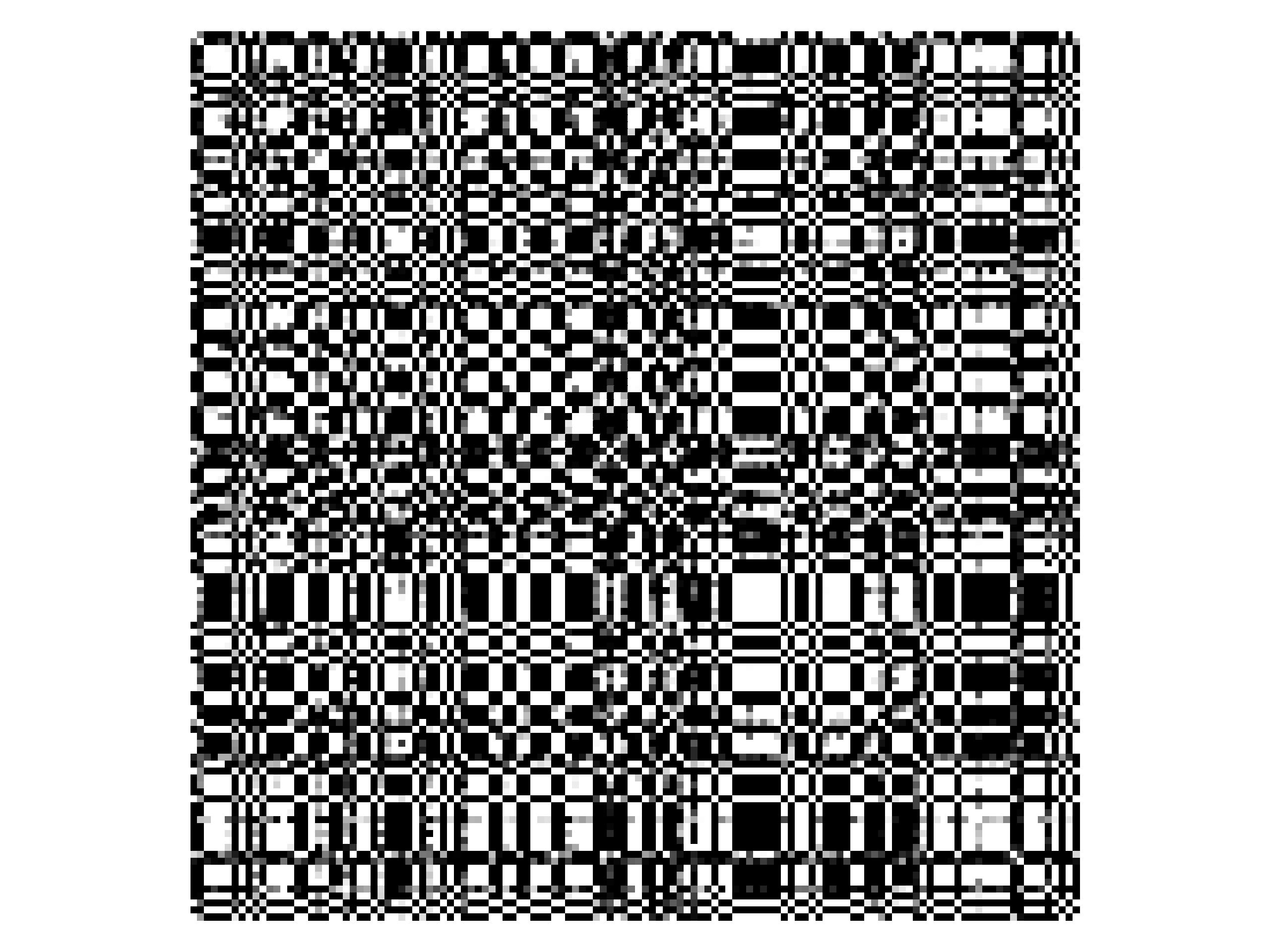}
    \includegraphics[width=0.3\linewidth]{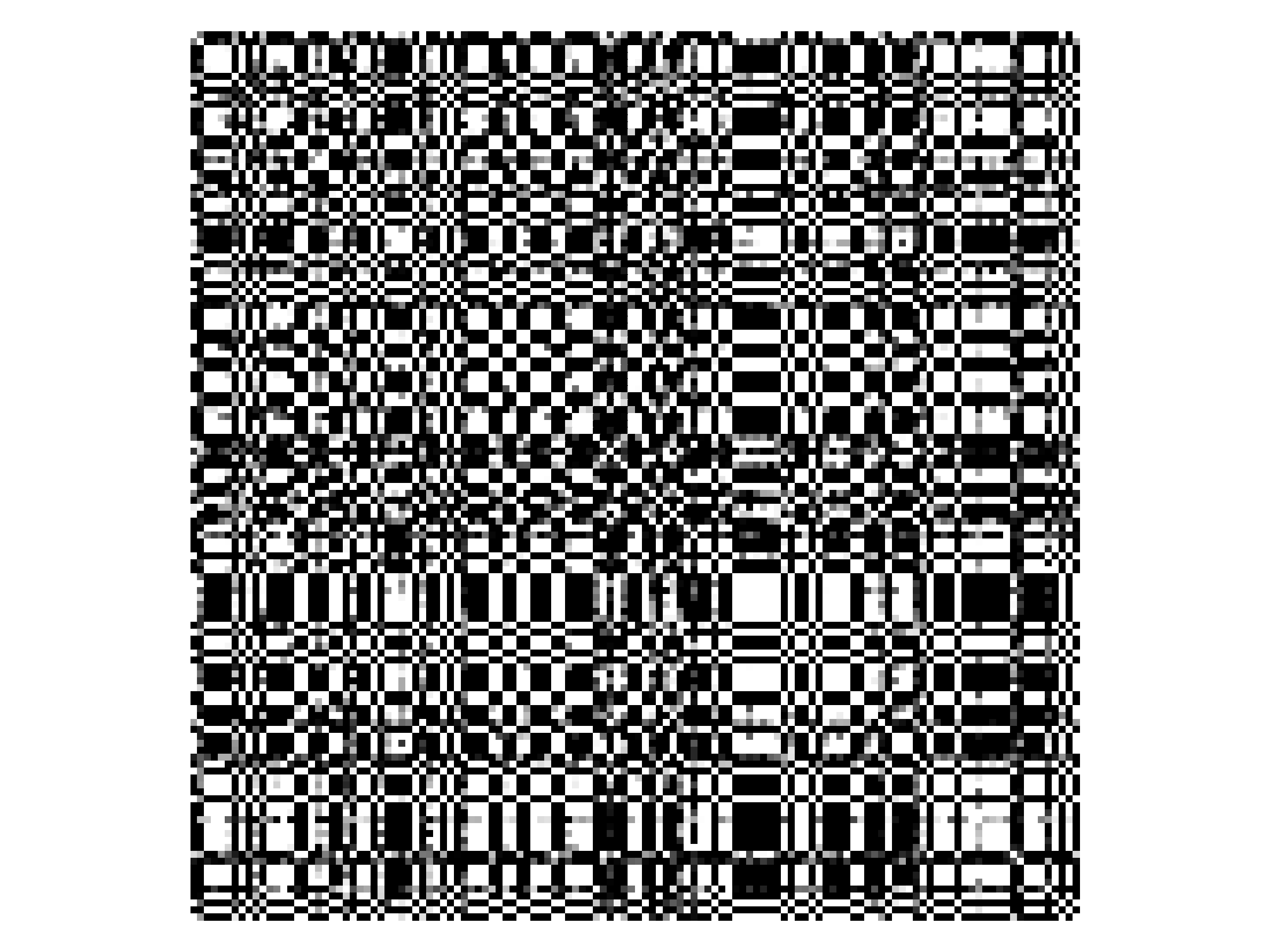}
    \includegraphics[width=0.3\linewidth]{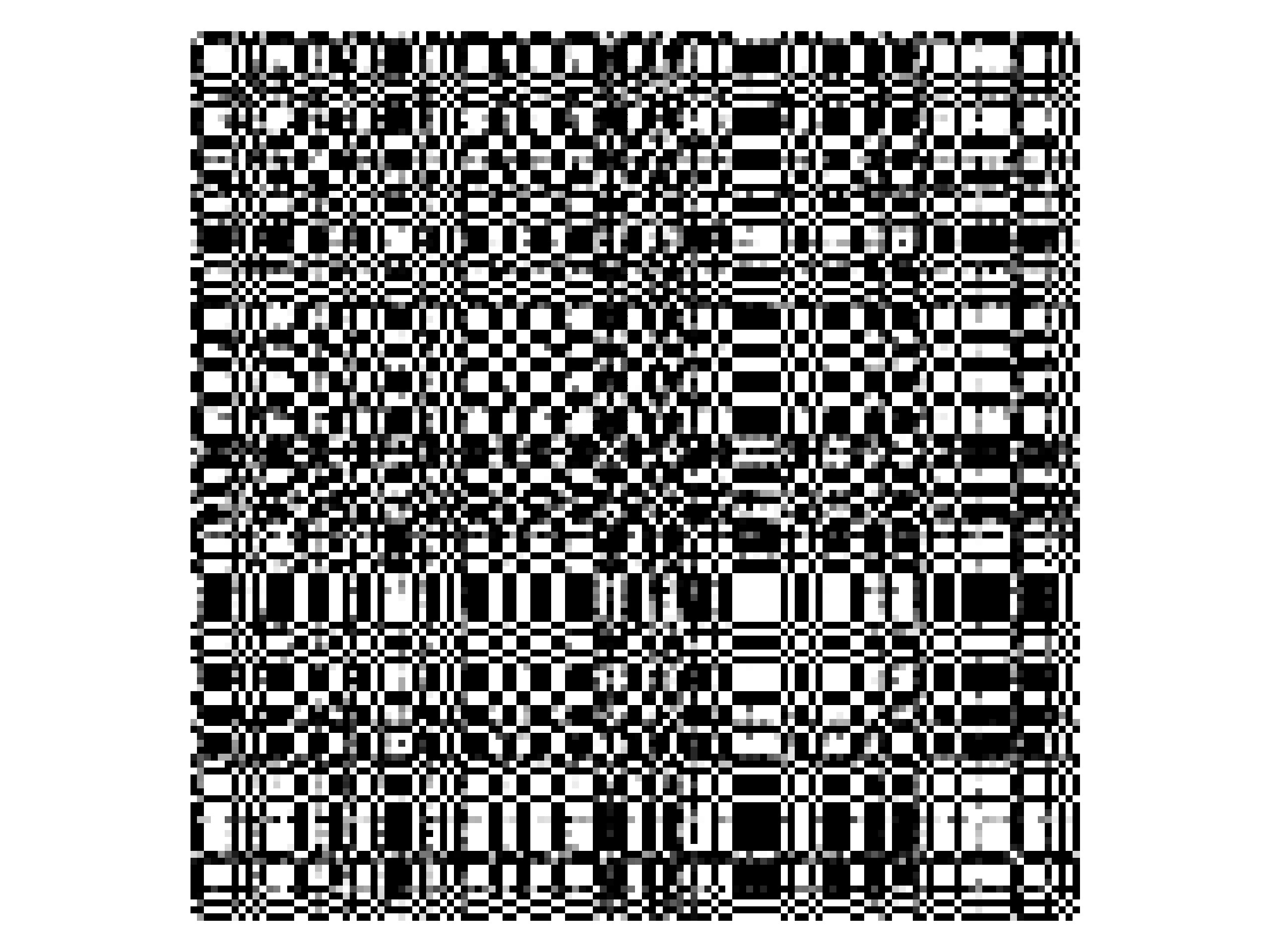}
  \end{subfigure}
  \begin{subfigure}{\linewidth}
  \captionsetup{labelformat=empty}
  \centering
  \rotatebox{90}{\quad  Epoch $150$}
    \includegraphics[width=0.3\linewidth]{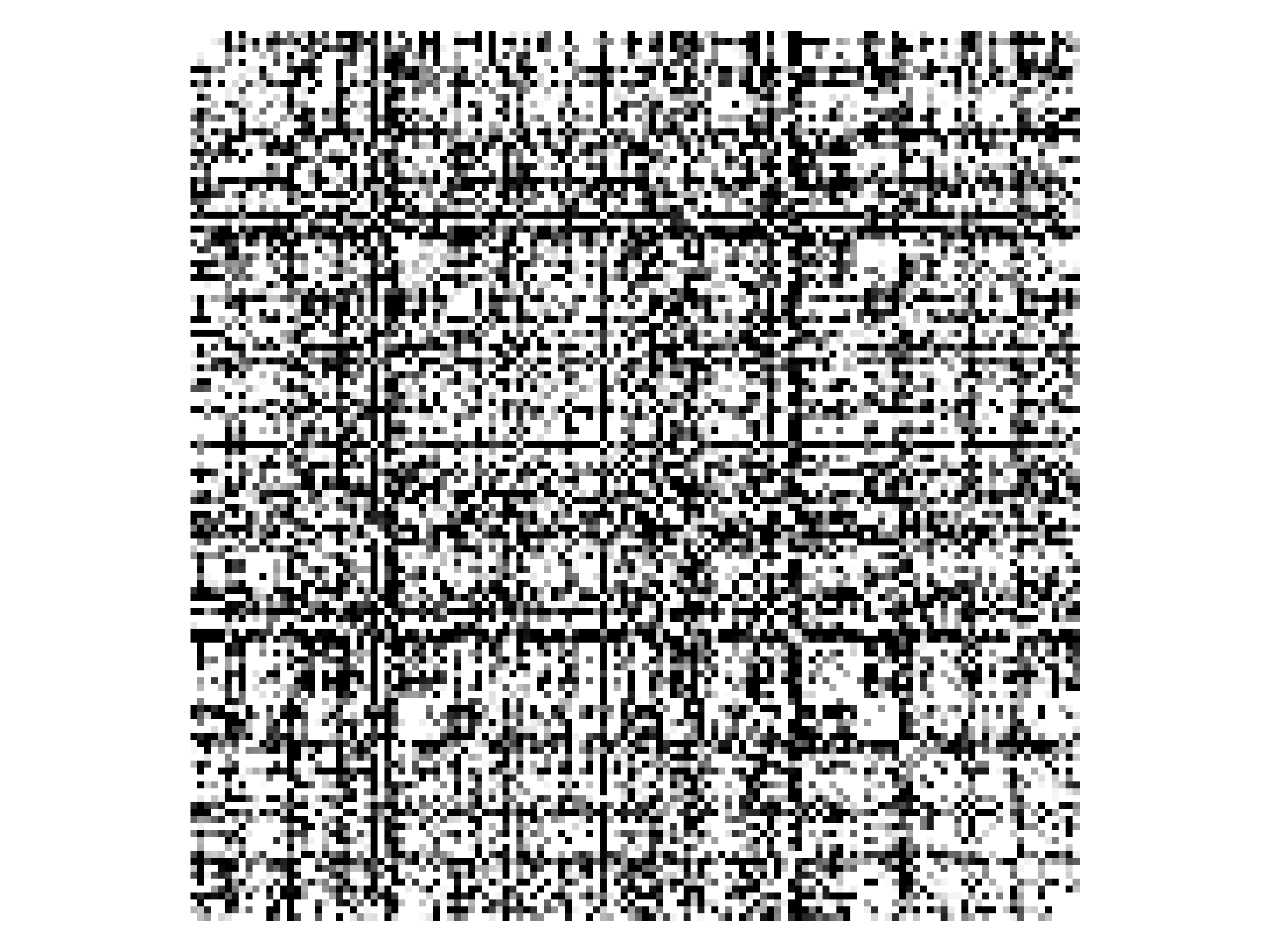}
    \includegraphics[width=0.3\linewidth]{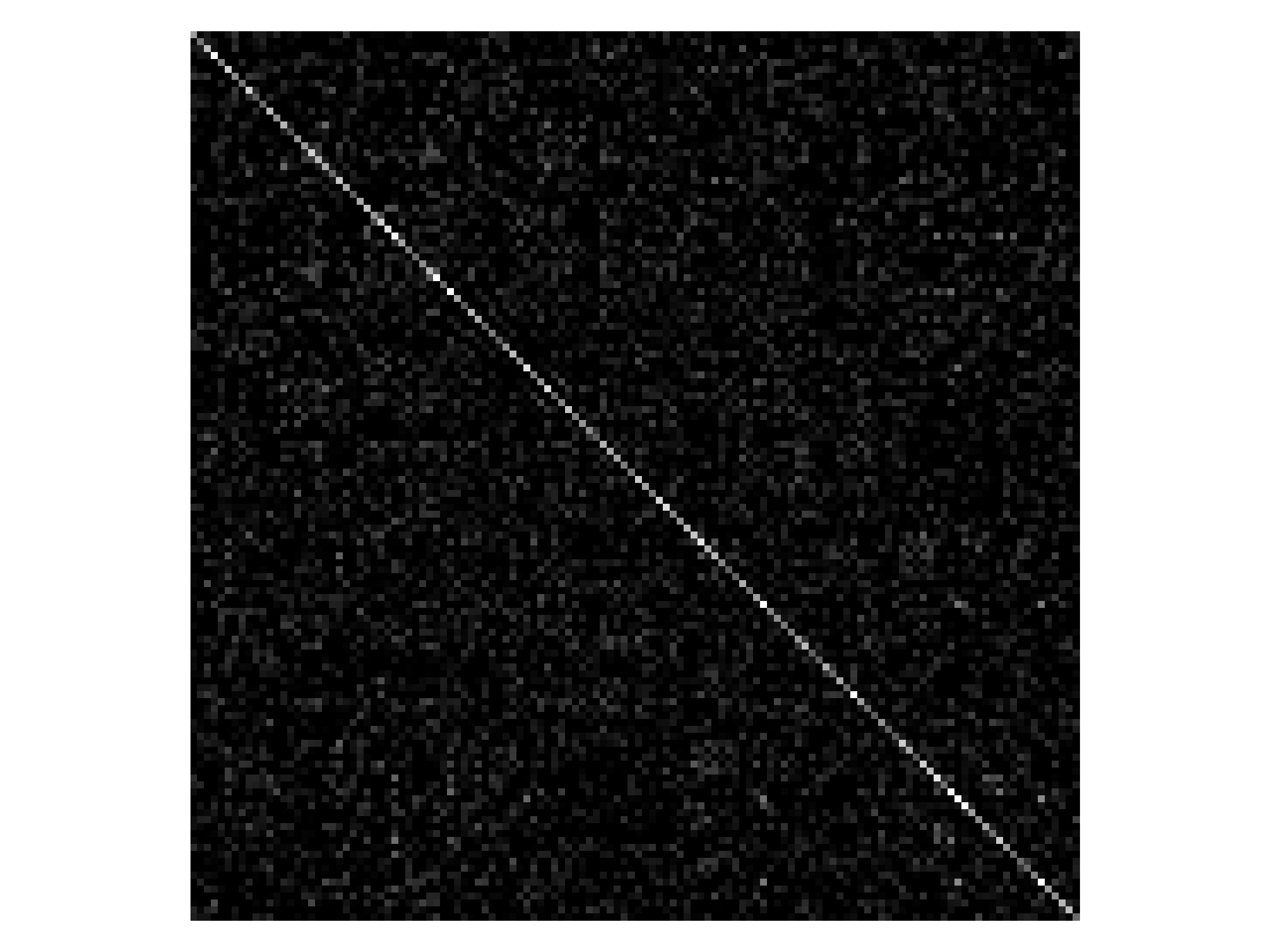}
    \includegraphics[width=0.3\linewidth]{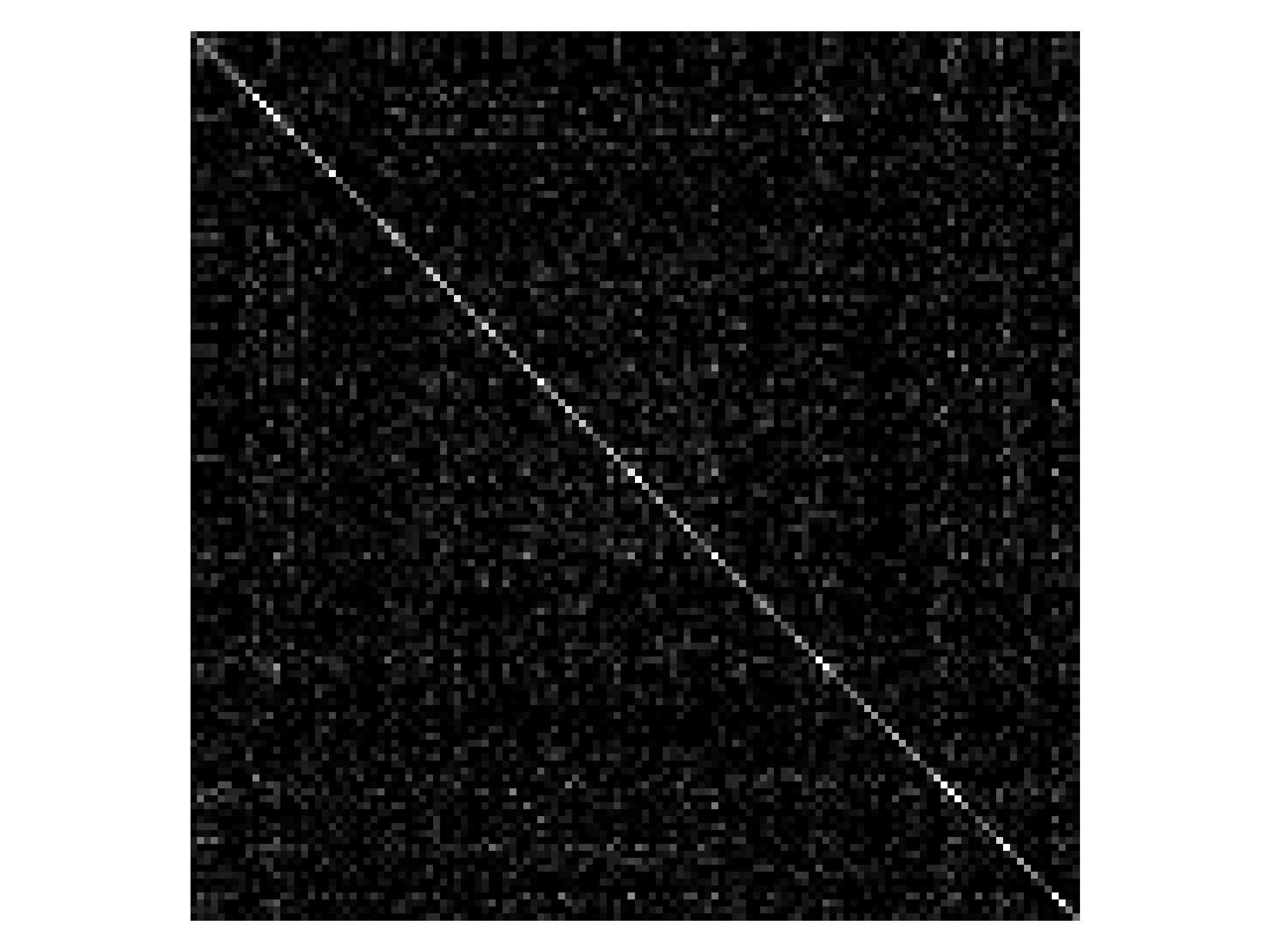}
  \caption{\hspace{25pt} BN \hspace{50pt} SWBN-KL \hspace{40pt} SWBN-Fro \hfill}
  \end{subfigure}
  \setlength{\belowcaptionskip}{-5pt}
  \caption{\small Comparison of heatmaps of correlation matrices. Columns, from left to right, represent heatmaps for the models with BN, SWBN-KL, and SWBN-Fro layers, respectively. Rows, from top to bottom, represent heatmaps of the corresponding models before and after training, respectively.}
  \label{fig:vgg_heatmap}
\end{figure}

To demonstrate the whitening effect of the SWBN algorithm, we feed in random $5000$ CIFAR-$10$ images to a trained VGG model, and extract its output of the last normalization layer, i.e. the features before being scaled and shifted by parameters $\vec{\gamma}$ and $\vec{\beta}$, respectively. Figure~\ref{fig:vgg_heatmap} shows the heatmap plots of the correlation matrices generated from the hidden features before training (i.e., epoch 0), and after $150$ epochs. For better visualization, we only show the correlation heatmaps for randomly selected $128$ features. In these plots, darker pixels represent smaller values in the correlation matrix. The plots in the first column show that \ac{BN} layers can not whiten the feature maps as batch normalized features remain highly correlated throughout training. The second and third columns of the plots show that the correlation matrices of the features after the SWBN layers are close to the identity matrix, indicating that the features are highly whitened. 


\subsubsection{Wall-clock Time Comparison}
\label{sec:time_compare}
To demonstrate the efficiency of \ac{SWBN}, we perform a series of experiments to measure the training time of \ac{SWBN} and \ac{ITERNORM}. We follow the same procedures as described in~\cite{huang2019iterative} to measure wall-clock time. We use TITAN Xp with Pytorch v$1.3$ and CUDA $10.1$ for the experiments. We define the input tensor $X \in \mathbf{R}^{h\times w \times d \times m}$ and the $3\times 3$ convolutional tensor $W\in \mathbf{R}^{3\times3\times d \times d}$, where $h=w=32$, $d$ and $m=64$ are height, width, number of channels and batch size, respectively. Pytorch implementations of \ac{SWBN} and \ac{ITERNORM} are used to run these experiments.  For each dimension $d\in \{256, 1024, 2048\}$, wall-clock time of one forward pass plus backward pass for a single layer is averaged over $100$ runs. Experimental results are summarized in Table~\ref{table_time}.

\begin{table}
\centering
\resizebox{\columnwidth}{!}{
\begin{tabular}{|c|c|c|c|}
\hline
\backslashbox{Layer Type}{Channels}     & d=256    & d=1024  & d=2048  \\ \hline
BatchNorm                             & 1.7ms  & 5.01ms & 11.44ms\\ \hline
SWBN-KL                              & 13.5ms & 72.5ms & 256.4ms \\ \hline
SWBN-Fro                             & 13.2ms & 72.3ms & 260.12ms \\ \hline
IterNorm                             & 15.5ms  & 111.8ms & 475.34ms\\ \hline
\end{tabular}}
\setlength{\belowcaptionskip}{-5pt}
\caption{\small Single layer wall-clock time (in ms), averaged over $100$ runs, for BatchNorm, \ac{ITERNORM} and \ac{SWBN}.}
\label{table_time}
\end{table}
\begin{table}
\centering
\resizebox{\columnwidth}{!}{
\begin{tabular}{|c|c|c|c|}
\hline
CIFAR-10 (1 GPU)     & SWBN-KL & SWBN-Fro & IterNorm     \\ \hline 
ResNetV2-164 & \bf{123}s    & 125s     & 187s         \\ \hline
WRN-40-10    & \bf{226}s   & 230s     & 299s         \\ \hline \hline
ImageNet (8 GPUs)     & SWBN-KL & SWBN-Fro & IterNorm     \\ \hline
ResNetV2-50  & \bf{27}min   & 31min    & 97min        \\ \hline
ResNetV2-101 & \bf{38}min   & 43min    & 184min \\ \hline
\end{tabular}}
\setlength{\belowcaptionskip}{-5pt}
\caption{\small Training time per epoch of IterNorm and \ac{SWBN} layers on CIFAR-10 and ImageNet datasets, averaged over $3$ runs.
}
\label{table_time_train}
\end{table}

To show the efficiency of \ac{SWBN} with popular \ac{DNN} models, we select the wide architecture Wide ResNet(WRN)~\cite{zagoruyko2016wide} and the deep architecture ResNetV2~\cite{he2016identity}. All \ac{BN} layers in these models are replaced with \ac{SWBN} and \ac{ITERNORM}, and the models are trained on CIFAR-$10$ dataset with images of size $32\times 32 \times 3$. Further, we do the same experiments with ResNetV2 on the ImageNet dataset with images of size $224\times 224\times 3$. The batch size is fixed to $128$ for CIFAR-$10$ and $256$ for ImageNet. On ImageNet, we use $8$ Tesla V$100$ GPUs for acceleration. Table~\ref{table_time_train} summarizes training time of one epoch for models with different whitening layers. As seen, \ac{SWBN} models are significantly faster than their \ac{ITERNORM} counterparts, especially for very deep CNNs with a large input size.


\subsection{Image Classification}
\label{exp:classification}
In this section, we evaluate the performance of SWBN on image classification benchmarks CIFAR-$10$, CIFAR-$100$ and ILSVRC-$2012$ (ImageNet). The performance of \ac{SWBN} is compared with that of the state-of-the-art whitening methods, such as \ac{DBN}~\cite{huang2018decorrelated} and \ac{ITERNORM}~\cite{huang2019iterative}. 

\subsubsection{CIFAR-$10$ and CIFAR-$100$}
\label{exp:cifar}
In this section, we provide \ac{SWBN}'s performance on CIFAR-$10$ and CIFAR-$100$ datasets using deep and wide CNNs. We select a deep model architecture, ResNetV2~\cite{he2016identity}, and a wide model architecture, Wide ResNet (WRN)~\cite{zagoruyko2016wide}. We use the same architecture as reported in~\cite{zagoruyko2016wide} and~\cite{he2016identity}, and replace all \ac{BN} layers with \ac{SWBN} or \ac{ITERNORM} layers. CIFAR-$100$ is a variant dataset of CIFAR-$10$, which has $60$K color images of size $32\times 32$: $50$K in the training set and $10$K in the test set. The task is to classify images into $100$ categories instead of $10$, making it more challenging than CIFAR-$10$ because there are fewer data samples for each category. Every experiment was repeated $10$ times with different random seeds. The mean test accuracies are reported in Tables~\ref{table_cifar10} and~\ref{table_cifar100}. We use the same training configuration, hyper-parameters and data augmentation setups as described in the original papers. The whitening step size $\alpha$ is set to $10^{-5}$ for all the experiments. Because no results on CIFAR-$100$ dataset are reported in~\cite{huang2019iterative}, we used the released code from~\cite{huang2019iterative} to run these experiments. We don't conduct additional experiments for \ac{DBN} because \ac{ITERNORM} is faster and has better performance~\cite{huang2019iterative}.


As shown in Table~\ref{table_cifar10}, for CIFAR-$10$ dataset models with SWBN layers generally outperform the ones with BN layers, and have similar performance as those with \ac{ITERNORM} layers. However, as shown in Table~\ref{table_cifar100}, for CIFAR-$100$ dataset, SWBN layers improve the generalization performance of these models. Surprisingly, \ac{ITERNORM} layers reduce the generalization performance of deep CNN models like ResNetV$2$ comparing with \ac{BN} layers. 


\begin{table}[htp]
\centering
\resizebox{\columnwidth}{!}{
\begin{tabular}{|c|c|c|c|c|c|}
\hline 
\multicolumn{1}{|c|}{Models} & \multicolumn{1}{c|}{BN Baseline} & \multicolumn{1}{c|}{DBN} & \multicolumn{1}{c|}{IterNorm} & \multicolumn{1}{c|}{SWBN-KL} & \multicolumn{1}{c|}{SWBN-Fro} \\ \hline \hline
ResNetV2-56  & 92.92  & N/A  & $93.11^*$   & \bf{93.4} & 93.23  \\ \hline
ResNetV2-164  & 94.54~\cite{he2016identity}  & N/A  & $94.45^*$   & \bf{94.68} & 94.59  \\ \hline
WRN-28-10  & 96.11~\cite{zagoruyko2016wide}  & 96.21~\cite{huang2018decorrelated}  & $96.19^*$ & \bf{96.23} & 96.18  \\ \hline
WRN-40-10  & 96.2~\cite{zagoruyko2016wide}   & 96.26~\cite{huang2018decorrelated}  & $96.23^*$ & \bf{96.39} & 96.35  \\ \hline
\end{tabular}}
\vspace{-5pt}
\caption{\small CIFAR-$10$ Results. All numbers for \ac{SWBN} models are averaged test accuracies (\%) of $10$ runs. Best average accuracies are highlighted in bold face, and N/A indicates that the accuracy is not reported in the referenced paper. $^*$ indicates results produced by the released code of~\cite{huang2019iterative}.}
\label{table_cifar10}

\vspace{7pt}

\resizebox{\columnwidth}{!}{
\begin{tabular}{|c|c|c|c|c|}
\hline
\multicolumn{1}{|c|}{Models} & \multicolumn{1}{c|}{BN Baseline} & \multicolumn{1}{c|}{IterNorm} & \multicolumn{1}{c|}{\skl} & \multicolumn{1}{c|}{\sfro} \\ \hline \hline
ResNetV2-56  & 73.01     & $72.1^*$   & \bf{73.12}  & 72.9 \\ \hline
ResNetV2-164  & 75.56     & $74.12^*$   & \bf{76.13}  & 76.02 \\ \hline
WRN-28-10  & 81.15~\cite{zagoruyko2016wide}     & $79.83^*$              & \bf{81.41}  & 81.31 \\ \hline
WRN-40-10  & 81.7~\cite{zagoruyko2016wide}      & $80.56^*$               & 81.7  & \bf{81.78} \\ \hline
\end{tabular}}
\vspace{-5pt}
\setlength{\belowcaptionskip}{-10pt}
\caption{\small CIFAR-$100$ Results. All numbers for SWBN models are average test accuracies (\%) of $10$ runs. Best average accuracies are highlighted in bold face. $^*$ indicates results produced by the released code of~\cite{huang2019iterative}. The \ac{BN} baseline accuracies reported for WRN in~\cite{zagoruyko2016wide} are the best single run results.}
\label{table_cifar100}
\end{table}

\subsubsection{ILSVRC-$2012$ (ImageNet)}

In this section, we compare \ac{SWBN} with \ac{ITERNORM} on ILSVRC-$2012$, a.k.a. ImageNet dataset. The dataset has $1.28$ million images for training and $50,000$ images for testing. The task is to classify an image into $1000$ classes. 

In~\cite{huang2019iterative}, to speed up the training of ResNet~\cite{he2016deep} with \ac{ITERNORM} on this dataset, the authors only replaced the first \ac{BN} layer with an \ac{ITERNORM} layer and added one \ac{ITERNORM} layer before the last linear layer, a total of two \ac{ITERNORM} layers used in their models. To make a fair comparison, we use the same setting, by replacing $2$ \ac{BN} layers with \ac{SWBN} layers at the exact locations in the model. The results are shown in the first row of Table~\ref{table:imagenet}.

To further test the scalability and performance improvement of SWBN for larger state-of-the-art models, we train two ResNeXt models~\cite{xie2017aggregated} with \ac{SWBN} layers. We employ the same configuration as defined in~\cite{xie2017aggregated}, and choose ResNeXt-$50$, $32$x$4$d and ResNeXt-$101$, $32$x$4$d, which have $\sim 25$M and $\sim 44$M parameters, respectively. Experimental configurations can be found in Appendix D. 

 Both top-$1$ and top-$5$ test accuracies are reported in Table~\ref{table:imagenet}. The test accuracies were evaluated on the single-cropped $224\times 224$ test images. As seen, the ResNeXt models that use SWBN layers outperform the ones with BN layers, both in top-1 and top-5 accuracies. This validates the scalability of the proposed SWBN layer for large networks and datasets. We were not able to perform identical experiments for \ac{ITERNORM} on ResNeXt due to high computational cost of \ac{ITERNORM}, as discussed in Section~\ref{sec:complexity} and indicated in Table~\ref{table_time} and~\ref{table_time_train}.
 
 \begin{table}[]
\centering
\resizebox{\columnwidth}{!}{
\begin{tabular}{|c||c|c||c|c||c|c||c|c|}
\hline
\multirow{2}{*}{Model} & \multicolumn{2}{c|}{BN} & \multicolumn{2}{c|}{IterNorm} & \multicolumn{2}{c|}{SWBN-KL} & \multicolumn{2}{c|}{SWBN-Fro} \\ \cline{2-9} 
                       & top-1      & top-5      & top-1         & top-5         & top-1         & top-5        & top-1         & top-5         \\ \hline \hline
ResNet-50        & 75.3~\cite{drn_website}       & 92.2~\cite{drn_website}       & \bf{77.09}~\cite{huang2019iterative}         & 93.53~\cite{huang2019iterative}         & 77.03         & \bf{93.61}        & 76.95         & 93.23         \\ \hline
ResNeXt-50        & 77.8       & N/A        & N/A           & N/A           & 78.1          & \bf{93.71}        & \bf{78.2}          & 93.68         \\ \hline
ResNeXt-101        & 78.8       & 94.4       & N/A           & N/A           & \bf{79.39}         & \bf{94.51}        & 79.27         & 94.48         \\ \hline
\end{tabular}
}
\vspace{-5pt}
\setlength{\belowcaptionskip}{-10pt}
\caption{\small Single run results of ILSVRC-$2012$. N/A indicates the accuracy is not reported in the referenced paper or not available for high computational cost.}
\label{table:imagenet}
\end{table}

\subsection{Few-shot Classification}
Few-shot classification aims to recognize unlabeled samples of newly observed classes given only one or a few labeled samples. Unknown data distributions of unseen classes and the scarce amount of labeled data make few-shot classification particularly difficult. In terminology of the few-shot classification, if the few-shot training (a.k.a support) dataset contains K labeled samples for each of C categories, the target few-shot task is called a C-way K-shot task. Metric learning, which stands for approaches designed to learn transferable data representations, is commonly used to tackle this task. Siamese networks~\cite{koch2015siamese}, matching networks~\cite{vinyals2016matching}, and prototypical networks~\cite{snell2017prototypical} are examples of metric learning models. Recently proposed cross attention networks~\cite{hou2019cross} shows the-state-of-the-art performance on benchmark datasets. Most of these approaches require training backbone networks for extracting representations from input data. We choose matching networks, prototypical networks, and cross attention networks with small backbone networks to compare the performance of \ac{BN}, \ac{SWBN}, and \ac{ITERNORM} layers. Experimental details are included in Appendix D. Table~\ref{table:fs_results} shows results on two few-shot classification benchmark datasets, namely mini-Imagenet and CIFAR-FS. We choose Resnet12 and Resnet20 as the backbone networks for mini-Imagenet and CIFAR-FS, respectively. As shown in Table~\ref{table:fs_results}, all the whitening layers outperform the \ac{BN} layer, and \sfro~is generally better than \ac{ITERNORM} while having lower memory consumption and 
better computational efficiency.

\begin{table}[]
\centering
\resizebox{\columnwidth}{!}{
\begin{tabular}{|c|c|c|c|c|c|c|c|c|c|c|}
\hline
\multirow{2}{*}{Dataset} &
  \multirow{2}{*}{Approach} &
  \multirow{2}{*}{Backbone} &
  \multicolumn{2}{c|}{BN} &
  \multicolumn{2}{c|}{SWBN-KL} &
  \multicolumn{2}{c|}{SWBN-Fro} &
  \multicolumn{2}{c|}{IterNorm} \\ \cline{4-11} 
 &
   &
   &
  5W1S &
  5W5S &
  5W1S &
  5W5S &
  5W1S &
  5W5S &
  5W1S &
  5W5S \\ \hline \hline
\multirow{3}{*}{\small \shortstack[c]{mini-\\ Imagenet}} &
  MN~\cite{vinyals2016matching} &
  \multirow{3}{*}{Resnet12} &
  57.37 &
  68.22 &
  \textbf{57.79} &
  \textbf{68.88} &
  57.55 &
  68.47 &
  57.49 &
  68.64 \\ \cline{2-2} \cline{4-11} 
 &
  PN~\cite{snell2017prototypical} &
   &
   55.29 &
   73.63 &
   55.44 &
   73.74 &
   \textbf{56.55} &
   \textbf{74.5} &
   55.66 &
   73.26\\ \cline{2-2} \cline{4-11} 
 &
  CAN~\cite{hou2019cross} &
   &
  62.58	&
  78.64&
  64.37&
  79.19&
  \textbf{64.97}&
  79.22&
  64.12 &
  \textbf{79.64} \\ \hline
\multirow{3}{*}{\small CIFAR-FS} &
  MN~\cite{vinyals2016matching} &
  \multirow{3}{*}{Resnet20} &
   61.28&
   72.8	&
   62.32&
   73.98&
   \textbf{62.73}&
   \textbf{74.26}&
   62.35&
   74.21 \\ \cline{2-2} \cline{4-11} 
 &
  PN~\cite{snell2017prototypical} &
   &
   55.73 &
   73.47 &
   56.76 &
   74.51 &
   \textbf{57.38} &
   \textbf{75.02} &
   56.61 &
   74.48\\ \cline{2-2} \cline{4-11} 
 &
  CAN~\cite{hou2019cross} &
   &
   65.28 &
   79.39 &
   65.71 &
   79.84 &
   \textbf{66.08} &
   80.45 &
   65.95 &
   \textbf{80.76}\\ \hline
\end{tabular}
}
\vspace{-5pt}
\setlength{\belowcaptionskip}{-10pt}
\caption{\small Results of mini-Imagenet and CIFAR-FS datasets. Best test accuracies are highlighted in bold face. The abbreviations in the Approach column stand for: matching network (MN), prototypical network (PN), and cross attention network (CAN). $c$W$k$S stands for $c$-way $k$-shot.}
\label{table:fs_results}
\end{table}

\section{Conclusions}
In this paper, we propose the \acf{SWBN} technique with two whitening criteria $C_{KL}$ and $C_{Fro}$. \ac{SWBN} is a new extension to \acf{BN}, which further whitens data in an online fashion. The proposed data whitening algorithm outperforms the newly proposed \ac{ITERNORM} in terms of computational efficiency. The SWBN layers accelerate training convergence of deep neural networks and enable them to have better generalization performance by incrementally whitening and rescaling activations. Ablation experiments demonstrate that SWBN is capable of efficiently whitening data in a stochastic way. The wall-clock time records show that \ac{SWBN} is more efficient than \ac{ITERNORM}. We also show the performance improvement by replacing \ac{BN} layers inside the state-of-the-art CNN models with \ac{SWBN} layers on CIFAR-$10$/$100$ and the ImageNet dataset, as well as few-shot classification benchmark datasets mini-Imagenet and CIFAR-FS.
%
%

\bibliographystyle{abbrv}
\bibliographystyle{splncs04} 
\bibliography{egbib}

\begin{thebibliography}{10}

\bibitem{Ba2016LayerN}
J.~Ba, R.~Kiros, and G.~E. Hinton.
\newblock Layer normalization.
\newblock {\em ArXiv}, abs/1607.06450, 2016.

\bibitem{bell1997independent}
A.~J. Bell and T.~J. Sejnowski.
\newblock The “independent components” of natural scenes are edge filters.
\newblock {\em Vision research}, 37(23):3327--3338, 1997.

\bibitem{bertinetto2018meta}
L.~Bertinetto, J.~F. Henriques, P.~H. Torr, and A.~Vedaldi.
\newblock Meta-learning with differentiable closed-form solvers.
\newblock {\em arXiv preprint arXiv:1805.08136}, 2018.

\bibitem{cardoso1996equivariant}
J.-F. Cardoso and B.~H. Laheld.
\newblock Equivariant adaptive source separation.
\newblock {\em IEEE Transactions on signal processing}, 44(12):3017--3030,
  1996.

\bibitem{chiley2019online}
V.~Chiley, I.~Sharapov, A.~Kosson, U.~Koster, R.~Reece, S.~Samaniego de~la
  Fuente, V.~Subbiah, and M.~James.
\newblock Online normalization for training neural networks.
\newblock {\em Advances in Neural Information Processing Systems},
  32:8433--8443, 2019.

\bibitem{Cogswell2015}
M.~Cogswell, F.~Ahmed, R.~B. Girshick, L.~Zitnick, and D.~Batra.
\newblock Reducing overfitting in deep networks by decorrelating
  representations.
\newblock In {\em 4th International Conference on Learning Representations,
  {ICLR} 2016, San Juan, Puerto Rico, May 2-4, 2016, Conference Track
  Proceedings}, 2016.

\bibitem{imagenet_cvpr09}
J.~Deng, W.~Dong, R.~Socher, L.-J. Li, K.~Li, and L.~Fei-Fei.
\newblock {ImageNet: A Large-Scale Hierarchical Image Database}.
\newblock In {\em IEEE/CVF Conference on Computer Vision and Pattern
  Recognition}, 2009.

\bibitem{Desjardins05Natural}
G.~Desjardins, K.~Simonyan, R.~Pascanu, and k.~kavukcuoglu.
\newblock Natural neural networks.
\newblock In C.~Cortes, N.~D. Lawrence, D.~D. Lee, M.~Sugiyama, and R.~Garnett,
  editors, {\em Advances in Neural Information Processing Systems 28}, pages
  2071--2079. Curran Associates, Inc., 2015.

\bibitem{eldar2003mmse}
Y.~C. Eldar and A.~V. Oppenheim.
\newblock Mmse whitening and subspace whitening.
\newblock {\em IEEE Transactions on Information Theory}, 49(7):1846--1851,
  2003.

\bibitem{drn_website}
K.~He, R.~S. Zhang, Xiangyu, and J.~Sun.
\newblock {\em Deep Residual Networks}.

\bibitem{he2016deep}
K.~He, X.~Zhang, S.~Ren, and J.~Sun.
\newblock Deep residual learning for image recognition.
\newblock In {\em Proceedings of the IEEE conference on computer vision and
  pattern recognition}, pages 770--778, 2016.

\bibitem{he2016identity}
K.~He, X.~Zhang, S.~Ren, and J.~Sun.
\newblock Identity mappings in deep residual networks.
\newblock In {\em European conference on computer vision}, pages 630--645.
  Springer, 2016.

\bibitem{hou2019cross}
R.~Hou, H.~Chang, M.~Bingpeng, S.~Shan, and X.~Chen.
\newblock Cross attention network for few-shot classification.
\newblock In {\em Advances in Neural Information Processing Systems}, pages
  4003--4014, 2019.

\bibitem{huang2018decorrelated}
L.~Huang, D.~Yang, B.~Lang, and J.~Deng.
\newblock Decorrelated batch normalization.
\newblock In {\em Proceedings of the IEEE Conference on Computer Vision and
  Pattern Recognition}, pages 791--800, 2018.

\bibitem{huang2019iterative}
L.~Huang, Y.~Zhou, F.~Zhu, Y.~Liu, and L.~Shao.
\newblock Iterative normalization: Beyond standardization towards efficient
  whitening.
\newblock In {\em Proceedings of the IEEE Conference on Computer Vision and
  Pattern Recognition}, pages 4874--4883, 2019.

\bibitem{Ioffe2017BatchRT}
S.~Ioffe.
\newblock Batch renormalization: Towards reducing minibatch dependence in
  batch-normalized models.
\newblock In I.~Guyon, U.~V. Luxburg, S.~Bengio, H.~Wallach, R.~Fergus,
  S.~Vishwanathan, and R.~Garnett, editors, {\em Advances in Neural Information
  Processing Systems 30}, pages 1945--1953. Curran Associates, Inc., 2017.

\bibitem{ioffe2015batch}
S.~Ioffe and C.~Szegedy.
\newblock Batch normalization: Accelerating deep network training by reducing
  internal covariate shift.
\newblock pages 448--456, 2015.

\bibitem{jolliffe2011principal}
I.~Jolliffe.
\newblock {\em Principal Component Analysis}.
\newblock Springer Berlin Heidelberg, Berlin, Heidelberg, 2011.

\bibitem{kessy2018optimal}
A.~Kessy, A.~Lewin, and K.~Strimmer.
\newblock Optimal whitening and decorrelation.
\newblock {\em The American Statistician}, 72(4):309--314, 2018.

\bibitem{koch2015siamese}
G.~Koch.
\newblock Siamese neural networks for one-shot image recognition.
\newblock 2015.

\bibitem{CIFAR10}
A.~Krizhevsky, V.~Nair, and G.~Hinton.
\newblock Cifar-10 (canadian institute for advanced research).

\bibitem{LeCun1998EfficientB}
Y.~LeCun, L.~Bottou, G.~B. Orr, and K.-R. M\"{u}ller.
\newblock Efficient backprop.
\newblock In {\em Neural Networks: Tricks of the Trade, This Book is an
  Outgrowth of a 1996 NIPS Workshop}, pages 9--50, London, UK, UK, 1998.
  Springer-Verlag.

\bibitem{Liao2016StreamingNT}
Q.~Liao, K.~Kawaguchi, and T.~A. Poggio.
\newblock Streaming normalization: Towards simpler and more
  biologically-plausible normalizations for online and recurrent learning.
\newblock {\em ArXiv}, abs/1610.06160, 2016.

\bibitem{Luo2017LearningDA}
P.~Luo.
\newblock Learning deep architectures via generalized whitened neural networks.
\newblock In D.~Precup and Y.~W. Teh, editors, {\em Proceedings of the 34th
  International Conference on Machine Learning}, volume~70 of {\em Proceedings
  of Machine Learning Research}, pages 2238--2246, International Convention
  Centre, Sydney, Australia, 06--11 Aug 2017. PMLR.

\bibitem{Raiko2012DeepLM}
T.~Raiko, H.~Valpola, and Y.~Lecun.
\newblock Deep learning made easier by linear transformations in perceptrons.
\newblock In N.~D. Lawrence and M.~Girolami, editors, {\em Proceedings of the
  Fifteenth International Conference on Artificial Intelligence and
  Statistics}, volume~22 of {\em Proceedings of Machine Learning Research},
  pages 924--932, La Palma, Canary Islands, 21--23 Apr 2012. PMLR.

\bibitem{Salimans2016WeightNA}
T.~Salimans and D.~P. Kingma.
\newblock Weight normalization: A simple reparameterization to accelerate
  training of deep neural networks.
\newblock In D.~D. Lee, M.~Sugiyama, U.~V. Luxburg, I.~Guyon, and R.~Garnett,
  editors, {\em Advances in Neural Information Processing Systems 29}, pages
  901--909. Curran Associates, Inc., 2016.

\bibitem{shen2020powernorm}
S.~Shen, Z.~Yao, A.~Gholami, M.~Mahoney, and K.~Keutzer.
\newblock Powernorm: Rethinking batch normalization in transformers.
\newblock In {\em International Conference on Machine Learning}, pages
  8741--8751. PMLR, 2020.

\bibitem{snell2017prototypical}
J.~Snell, K.~Swersky, and R.~Zemel.
\newblock Prototypical networks for few-shot learning.
\newblock In {\em Advances in neural information processing systems}, pages
  4077--4087, 2017.

\bibitem{srivastava2014dropout}
N.~Srivastava, G.~Hinton, A.~Krizhevsky, I.~Sutskever, and R.~Salakhutdinov.
\newblock Dropout: a simple way to prevent neural networks from overfitting.
\newblock {\em The Journal of Machine Learning Research}, 15(1):1929--1958,
  2014.

\bibitem{vinyals2016matching}
O.~Vinyals, C.~Blundell, T.~Lillicrap, D.~Wierstra, et~al.
\newblock Matching networks for one shot learning.
\newblock In {\em Advances in neural information processing systems}, pages
  3630--3638, 2016.

\bibitem{wang2019backpropagation}
W.~Wang, Z.~Dang, Y.~Hu, P.~Fua, and M.~Salzmann.
\newblock Backpropagation-friendly eigendecomposition.
\newblock In {\em Advances in Neural Information Processing Systems}, pages
  3162--3170, 2019.

\bibitem{GroupNorm2018}
Y.~Wu and K.~He.
\newblock Group normalization.
\newblock In {\em Proceedings of the European Conference on Computer Vision
  (ECCV)}, pages 3--19, 2018.

\bibitem{xie2017aggregated}
S.~Xie, R.~Girshick, P.~Doll{\'a}r, Z.~Tu, and K.~He.
\newblock Aggregated residual transformations for deep neural networks.
\newblock In {\em Proceedings of the IEEE conference on computer vision and
  pattern recognition}, pages 1492--1500, 2017.

\bibitem{Xiong2016}
W.~Xiong, B.~Du, L.~Zhang, R.~Hu, and D.~Tao.
\newblock Regularizing deep convolutional neural networks with a structured
  decorrelation constraint.
\newblock {\em 2016 IEEE 16th International Conference on Data Mining (ICDM)},
  pages 519--528, 2016.

\bibitem{yoshida2017spectral}
Y.~Yoshida and T.~Miyato.
\newblock Spectral norm regularization for improving the generalizability of
  deep learning.
\newblock {\em arXiv preprint arXiv:1705.10941}, 2017.

\bibitem{zagoruyko2016wide}
S.~Zagoruyko and N.~Komodakis.
\newblock Wide residual networks.
\newblock {\em arXiv preprint arXiv:1605.07146}, 2016.

\end{thebibliography}

\end{document}


\appendix
\section*{Appendix}
\addcontentsline{toc}{section}{Appendix}
\renewcommand{\thesubsection}{\Alph{subsection}}
\renewcommand{\thefigure}{A. \arabic{figure}}
\setcounter{figure}{0}
\subsection{Model Architecture of VGG network}
\label{app:vgg_detail}
Figure~\ref{app:fig:vgg} shows the VGG architecture we used in our experiments. 

\tikzstyle{block} = [rectangle, draw, fill=gray!20, 
    text width=10em, text centered, rounded corners, minimum height=2em,node distance=0.75cm]
\tikzstyle{plus} = [rectangle, draw, fill=gray!70, 
    text width=2em, text centered, rounded corners, minimum height=2em]
\tikzstyle{plus2} = [rectangle, draw, fill=gray!10, 
    text width=2em, text centered, rounded corners, minimum height=2em]
\tikzstyle{resblock1_detail} = [rectangle, draw, fill=gray!70, 
    text width=10em, text centered, rounded corners, minimum height=2em, node distance=0.75cm]
\tikzstyle{resblock2_detail} = [rectangle, draw, fill=gray!10, 
    text width=10em, text centered, rounded corners, minimum height=2em, node distance=0.75cm]
\tikzstyle{resblock1} = [rectangle, draw, fill=gray!70, 
    text width=10em, text centered, rounded corners, minimum height=2em]
 \tikzstyle{resblock2} = [rectangle, draw, fill=gray!10, 
    text width=10em, text centered, rounded corners, minimum height=2em]
\tikzstyle{deepblock} = [rectangle, draw, fill=gray!40, 
    text width=10em, text centered, rounded corners, minimum height=3em,node distance=.9cm]

\begin{figure}[scale=0.3 every node/.style={transform shape}]
\centering
\resizebox{0.5\textwidth}{!}{%
\begin{tikzpicture}[node distance = 1cm, auto]
  \node [block] (input) {Input};
  \node[above of=input](aux_in){};
  \node [block, below of=input] (conv_1) {Conv(64, $3\times3$, 1) };
  \node [block, below of=conv_1] (bn_act_1) {BN/SWBN+ReLU};
 \node [block, below of=bn_act_1] (conv_2) {Conv(64, $3\times3$, 1) };
  \node [block, below of=conv_2] (bn_act_2) {BN/SWBN+ReLU};
 \node[block, below of=bn_act_2](max_pool1){Max pooling $2\times 2$};
   \node [block, below of=max_pool1] (conv_3) {Conv(128, $3\times3$, 1) };
   \node [block, below of=conv_3] (bn_act_3) {BN/SWBN+ReLU};
 \node [block, below of=bn_act_3] (conv_4) {Conv(128, $3\times3$, 1) };
  \node [block, below of=conv_4] (bn_act_4) {BN/SWBN+ReLU};
 \node[block, below of=bn_act_4](max_pool2){Max pooling $2\times 2$};
   \node [block, below of=max_pool2] (conv_5) {Conv(256, $3\times3$, 1) };
  \node [block, below of=conv_5] (bn_act_5) {BN/SWBN+ReLU};
 \node [block, below of=bn_act_5] (conv_6) {Conv(256, $3\times3$, 1) };
  \node [block, below of=conv_6] (bn_act_6) {BN/SWBN+ReLU};
 \node[block, below of=bn_act_6](max_pool3){Max pooling($2\times 2$)};
   \node[ right of=input, node distance=4.5cm](aux_in2){};
   \node [block, below of=aux_in2] (conv_7) {Conv(512, $3\times3$, 1) };
   \node [block, below of=conv_7] (bn_act_7) {BN/SWBN+ReLU};
 \node [block, below of=bn_act_7] (conv_8) {Conv(512, $3\times3$, 1) };
  \node [block, below of=conv_8] (bn_act_8) {BN/SWBN+ReLU};
 \node[block, below of=bn_act_8](max_pool4){Max pooling($2\times 2$)};
   \node [block, below of=max_pool4] (conv_9) {Conv(512, $3\times3$, 1) };
   \node [block, below of=conv_9] (bn_act_9) {BN/SWBN+ReLU};
   \node[block, below of=bn_act_9](max_pool5){Average pooling($2\times 2$)};
   \node [block, below of=max_pool5] (fc1) {Dense(10)};
   \node [block, below of=fc1] (softmax) {Softmax};
   \node[below of=softmax](aux_out){};
   \draw[->,black](max_pool3.south)|- ++ (+2.2cm,-0.8cm)|- (conv_7);
 \draw[->,black](softmax)--(aux_out.mid);
 \draw[->,black](aux_in.mid)--(input);
   
\end{tikzpicture}}
\caption{VGG architecture. Conv($c$, $k\times k$, $s$), represents a convolutional layer with $c$ channels, kernel of size $k\times k$ and stride of $s$. Dense($h$) represents fully connected layer with $h$ neurons. BN/SWBN represents a normalization layer. In our experiments, we used BN for baseline models and SWBN for our proposed architectures. ReLU represent rectified linear unit layer. Max(Average) pooling($d\times d$) represents max(average) pooling layer with $d\times d$ kernel.}\label{app:fig:vgg}
\end{figure}

\subsection{Model Architecture of ResNet network}
\label{app:resnet_detail}
\renewcommand{\thefigure}{B. \arabic{figure}}
\setcounter{figure}{0}
Figures~\ref{app:fig:resnet},~\ref{app:fig:resnet1}, and~\ref{app:fig:resnet2} show the ResNet architecture, we used in our experiments.
\begin{figure}[h]
\centering
\resizebox{0.5\textwidth}{!}{%
\begin{tikzpicture}
    \node [block] (input) {Input};
    \node[above of=input](aux_in){};
    \node [block, below of=input] (conv_1) {Conv(16, $3\times3$, 1) };
    \node [block, below of=conv_1] (bn_act_1) {BN/SWBN+ReLU};
    
    \node[deepblock, below of=bn_act_1](deepblock){Residual blocks};
    \node[block, below of=deepblock,node distance=.9cm](bn_act_2){BN/SWBN+ReLU};
    \node[block, below of=bn_act_2](avr_pooling){Average pooling};
    \node[block, below of=avr_pooling](dense){Dense(10)};
    \node[block, below of=dense](softmax){Softmax};
    \node[below of=softmax](aux_out){};
    \node[resblock1, right of=conv_1, node distance=5cm](res_1){Residual block(5, 16) \\ type one};
    \node[resblock2, below of=res_1](res_2){Residual block(1, 32) \\type two};
    \node[resblock1, below of=res_2](res_3){Residual block(4, 32) \\type one};
    \node[resblock2, below of=res_3](res_4){Residual block(1, 64)\\ type two};
    \node[resblock1, below of=res_4](res_5){Residual block(4, 64) \\type one};
  
   \draw[dashed] (deepblock.north east) -- (res_1.north west);
   \draw[dashed] (deepblock.south east) -- (res_5.south west);
   \draw[->,black](aux_in.mid)--(input);
   \draw[->,black](softmax)--(aux_out.mid);
 
\end{tikzpicture}}
\caption{ResNet architecture. Conv($c$, $k\times k$, $s$) represents a convolutional layer with $c$ channels, kernel of size $k\times k$ and stride of $s$. Residual block $(r, c)$, represents $r$ repitions of a residual unit, with $c$ channels in its convolutional layers. Dense($h$) represents fully connected layer with $h$ neurons. BN/SWBN represents a normalization layer. In our experiments, we used BN for baseline models and SWBN for our proposed architectures. ReLU represent rectified linear unit layer. }\label{app:fig:resnet}
\end{figure}

\begin{figure}[h]
\centering
\resizebox{0.4\textwidth}{!}{%
\begin{tikzpicture}
    \node[resblock1_detail](res_t1_conv1){Conv($c$, $3\times 3$, 1)};
    \node[resblock1_detail, below of=res_t1_conv1](res_t1_bn_act){BN/SWBN+ReLU};
    \node[resblock1_detail, below of=res_t1_bn_act](res_t1_conv2){Conv($c$, $1\times 1$, 1)};
    \node[resblock1_detail, below of =res_t1_conv2](res_t1_bn){BN/SWBN};
    \node[plus, below of =res_t1_bn](plus){+};
    \node[resblock1_detail,below of=plus, node distance=.9cm](res_t1_act){ReLU};
    \node[below of =res_t1_act](res_t1_aux_out){};
    \node[above of =res_t1_conv1](res_t1_aux_in){};

   \draw[->,black, looseness=2.5](res_t1_aux_in.south)to[out=0cm, in=0cm](plus);
   \draw[->, black](res_t1_act)--(res_t1_aux_out.mid);
   \draw[->,black](res_t1_bn)--(plus);
    \draw[->,black](plus)--(res_t1_act);
   \draw[->, black](res_t1_aux_in.mid)--(res_t1_conv1);
   
\end{tikzpicture}}
\caption{Residual unit-type one. Conv($c$, $k\times k$, $s$) represents a convolutional layer with $c$ channels, kernel of size $k\times k$ and stride of $s$. }\label{app:fig:resnet1}
\end{figure}

\begin{figure}
\centering
\resizebox{0.5\textwidth}{!}{%
\begin{tikzpicture}
    \node[resblock2_detail](res_t1_conv1){Conv(c, $3\times 3$, 2)};
    \node[resblock2_detail, below of=res_t1_conv1](res_t1_bn_act){BN/SWBN+ReLU};
    \node[resblock2_detail, below of=res_t1_bn_act](res_t1_conv2){Conv(c, $1\times 1$, 1)};
    \node[resblock2_detail, below of =res_t1_conv2](res_t1_bn){BN/SWBN};
    \node[resblock2_detail, right of=res_t1_bn_act, node distance=4cm](res_t1_conv3){Conv(c, $1\times 1$, 2)};
    \node[resblock2_detail, below of =res_t1_conv3](res_t1_bn2){BN};
    \node[plus2, below of =res_t1_bn](plus){+};
    \node[resblock2_detail,below of=plus, node distance=.9cm](res_t1_act){ReLU};
    \node[below of =res_t1_act](res_t1_aux_out){};
    \node[above of =res_t1_conv1](res_t1_aux_in){};

   \draw[->,black](res_t1_aux_in.south)-|(res_t1_conv3.north);
   \draw[->,black](res_t1_bn2)|-(plus);
   \draw[->, black](res_t1_act)--(res_t1_aux_out.mid);
   \draw[->,black](res_t1_bn)--(plus);
    \draw[->,black](plus)--(res_t1_act);
   \draw[->, black](res_t1_aux_in.mid)--(res_t1_conv1);
   
\end{tikzpicture}}
\caption{Residual unit-type two. This unit is used when changing feature map size and depth. Conv($c$, $k\times k$, $s$) represents a convolutional layer with $c$ channels, kernel of size $k\times k$ and stride of $s$. We keep the BN layer in the right side branch for both base line models and the SWBN models.  }\label{app:fig:resnet2}
\end{figure}

\subsection {More Details on Whitening Criteria}
\label{app:proofs}
\subsubsection { $C_{KL}$}
As shown in~\cite{boyd2004convex}, $C_{KL}$ is convex if $W\Sigma_{\vec{x}}W^T$ is positive definite. This however is a very strict condition and generally does not hold. However, it is proved in \cite{cardoso1996equivariant} that by iteratively updating $W$ using the Eq.~(3), $W$ will converge to a whitening matrix. The update rule for this criterion was derived based on "relative gradient" introduced in \cite{cardoso1996equivariant} as against the actual gradient.

The reason why we didn't use the actual gradient of $C_{KL}$ for updating $W$ is due to its high computational cost. The gradient of $C_{KL}$ with respect to $W$ is derived as \footnote{For the matrix gradient formulas in steps ($1$) and ($2$), please refer to~\cite{petersen2008matrix}.}:

\begin{small}
\begin{flalign}
\frac{\partial C_{KL}}{\partial W} &= \frac{1}{2}(\frac{\partial }{\partial W}tr(W\Sigma_{\vec{x}}W^T)-\frac{\partial }{\partial W}\ln\,\det(W\Sigma_{\vec{x}}W^T)) \\
&= \frac{1}{2} (\frac{\partial }{\partial W}tr(W\Sigma_{\vec{x}}W^T) - \frac{\partial }{\partial W}(\ln\,\det(W) \notag \\ 
& \quad \, +\ln\,\det(\Sigma_{\vec{x}}) +\ln\,\det(W^T))) \\
&= \frac{1}{2} (2W\Sigma_{\vec{x}} + 2(W^{-1})^T) \\
&= W\Sigma_{\vec{x}} + (W^{-1})^T 
\end{flalign}
\end{small}

As we can see, this gradient formula involves matrix inversion, which is usually computationally expensive and numerically unstable. The formula using relative gradient in Eq.~(3) nicely avoids these problems.

\subsubsection{$C_{Fro}$}

The update formula for $W$ using $C_{Fro}$ can be directly derived by computing the gradient matrix \footnote{For the matrix gradient formulas in steps ($3$) and ($4$), please refer to \cite{petersen2008matrix}.}:

\begin{small}
\begin{flalign}
\frac{\partial C_{Fro}}{\partial W} &= \frac{\partial }{\partial W} \frac{1}{2}||I - W\Sigma_{\vec{x}}W^T||_{Fro} \\
&= \frac{\partial }{\partial W}\frac{1}{2}[tr((I - W\Sigma_{\vec{x}}W^T)^T(I - W\Sigma_{\vec{x}}W^T)]^{\frac{1}{2}}  \\
&= \frac{1}{4||I - W\Sigma_{\vec{x}}W^T ||_{Fro}} (-2\frac{\partial}{\partial W}tr(W\Sigma_{\vec{x}}W^T) + \notag\\
& \quad\,\, \frac{\partial}{\partial W}tr(W\Sigma_{\vec{x}}W^TW\Sigma_{\vec{x}}W^T)) \\
&= \frac{1}{4||I - W\Sigma_{\vec{x}}W^T ||_{Fro}} (4W\Sigma_{\vec{x}}W^TW\Sigma_{\vec{x}}-4W\Sigma_{\vec{x}}) \\
&= \frac{1}{||I - W\Sigma_{\vec{x}}W^T ||_{Fro}} (W\Sigma_{\vec{x}}W^T-I)W\Sigma_{\vec{x}} 
\end{flalign}
\end{small}

Clearly, $C_{Fro}$ is a non-convex function. For example, in a one dimensional case, with $\Sigma_{\vec{x}}=1$ and $W$ as a scalar variable $w$, $C_{Fro} = \sqrt{(1-w^2)^2} = |1-w^2|$, which has two global minima at $w=1$ and $w=-1$. 

\subsection{Derivation of Back-propagation Algorithm}

Let $X\in \mathbb{R}^{d\times n}$ and $\hat{X}\in \mathbb{R}^{d\times n}$ be its input and output of an SWBN layer of a DNN, respectively. Given the loss function $L$, the gradient matrix $\frac{\partial L}{\partial \hat{X}}$ of $L$, and $\vec{\mu}$, $\vec{v}$, $X^S$, $X^W$ from Algorithm $1$, we compute the gradient matrix $\frac{\partial L}{\partial X}$ by the chain rule. Without the loss of generality, we derive the formula of the partial derivative $\frac{\partial L}{\partial \hat{X}_{kl}}$ as:

\small{
\begin{align*}
\frac{\partial L}{\partial X_{kl}} &= \Sigma_{i=1}^d \frac{\partial L}{\partial \hat{X}_{il}}\frac{\partial \hat{X}_{il}}{\partial X_{kl}} \\ 
\frac{\partial \hat{X}_{il}}{\partial X_{kl}} &= \frac{\partial}{\partial X_{kl}}(\gamma_i X^W_{il} + \beta_i)  = \gamma_i \frac{\partial X^W_{il}}{\partial X_{kl}} \\
\frac{\partial X^W_{il}}{\partial X_{kl}} &= \frac{\partial }{\partial X_{kl}}\Sigma_{t=1}^d W_{it}X^S_{tl}=W_{ik}\frac{\partial X^S_{kl}}{\partial X_{kl}} \\
\frac{\partial X^S_{kl}}{\partial X_{kl}} &= \frac{\partial}{\partial X_{kl}} \frac{X_{kl}-\mu_k(X_{kl})}{\sqrt{v_k(X_{kl})+\epsilon}}\\
  &=  \frac{1}{\sqrt{v_k+\epsilon}} + \frac{\partial X^{S}_{kl}}{\partial v_k}\frac{\partial v_k}{\partial X_{kl}}+ \frac{\partial X^{S}_{kl}}{\partial \mu_k}\frac{\partial \mu_k}{\partial X_{kl}}\\
  \frac{\partial X^S_{kl}}{\partial v_k}&=-\frac{1}{2}(X_{kl}-\mu_k)(v_k+\epsilon)^{-\frac{2}{3}}\\
\frac{\partial X^S_{kl}}{\partial \mu_k} &= -\frac{1}{\sqrt{v_k + \epsilon}}\\
\frac{\partial v_k}{\partial X_{kl}}&=\frac{2}{n}(X_{kl}-\mu_k)\\
\frac{\partial \mu_k}{\partial X_{kl}}&=	1/n
\end{align*}
}
\subsection{SWBN with Different Model Architectures}
\label{sub:model_arch_sess}

We start with a fully-connected network on MNIST~\cite{lecun1998gradient} dataset. We then move to popular CNN architectures i.e., VGG~\cite{simonyan2014very} and ResNetV$1$~\cite{he2016deep} on CIFAR-$10$~\cite{CIFAR10}. We used the MNIST dataset to verify the performance of \acp{SWBN} in fully-connected neural networks. The MNIST dataset consists of $60,000$ images for training and $10,000$ images for testing where each image is a $28\times28$ pixel gray-scale image of a hand-written digit ranging from $0$ to $9$. The task is to predict the label of an input image. We use a network of $3$ fully-connected hidden layers and one fully-connected output layer. Each hidden layer has $1024$ neurons with ReLU activation functions, and the output layer has $10$ neurons with the softmax activation function. Normalization layers BN, \skl, or \sfro, are added before ReLU functions. The network was trained for $100$ epochs with softmax cross-entropy loss, with learning rate halved after every $20$ epochs.

For CNN models, we used CIFAR-$10$ dataset to make performance comparison on VGG networks and ResNetV$1$ with difference normalization layers. CIFAR-$10$ has $60,000$, $32\times 32$ pixels color images, $50,000$ in training set and $10,000$ in test set. The task is to classify images into $10$ categories. Training configuration was the same as described before, except that the network was trained for $150$ epochs and the learning rate was halved after every $30$ epoch. The chosen VGG network has $9$ convolutional layers and $1$ fully-connected layer. We placed normalization layers as explained in section~\ref{sub:ablation} for the fully-connected networks. The whitening step size $\alpha$ is set to $10^{-5}$. For the ResNet experiment, we selected ResNetV$1$ with depth of $32$ described in \cite{he2016deep}. When creating \ac{SWBN} models, we replaced all \ac{BN} layers in the original model with \ac{SWBN} layers except the ones in linear projection residual shortcuts, which were used to match dimension changes. All the model architecture details are included in the Appendix-sections~\ref{app:vgg_detail} and~\ref{app:resnet_detail}

\begin{figure}
\centering
\subfigure{
\includegraphics[width=0.45\linewidth]{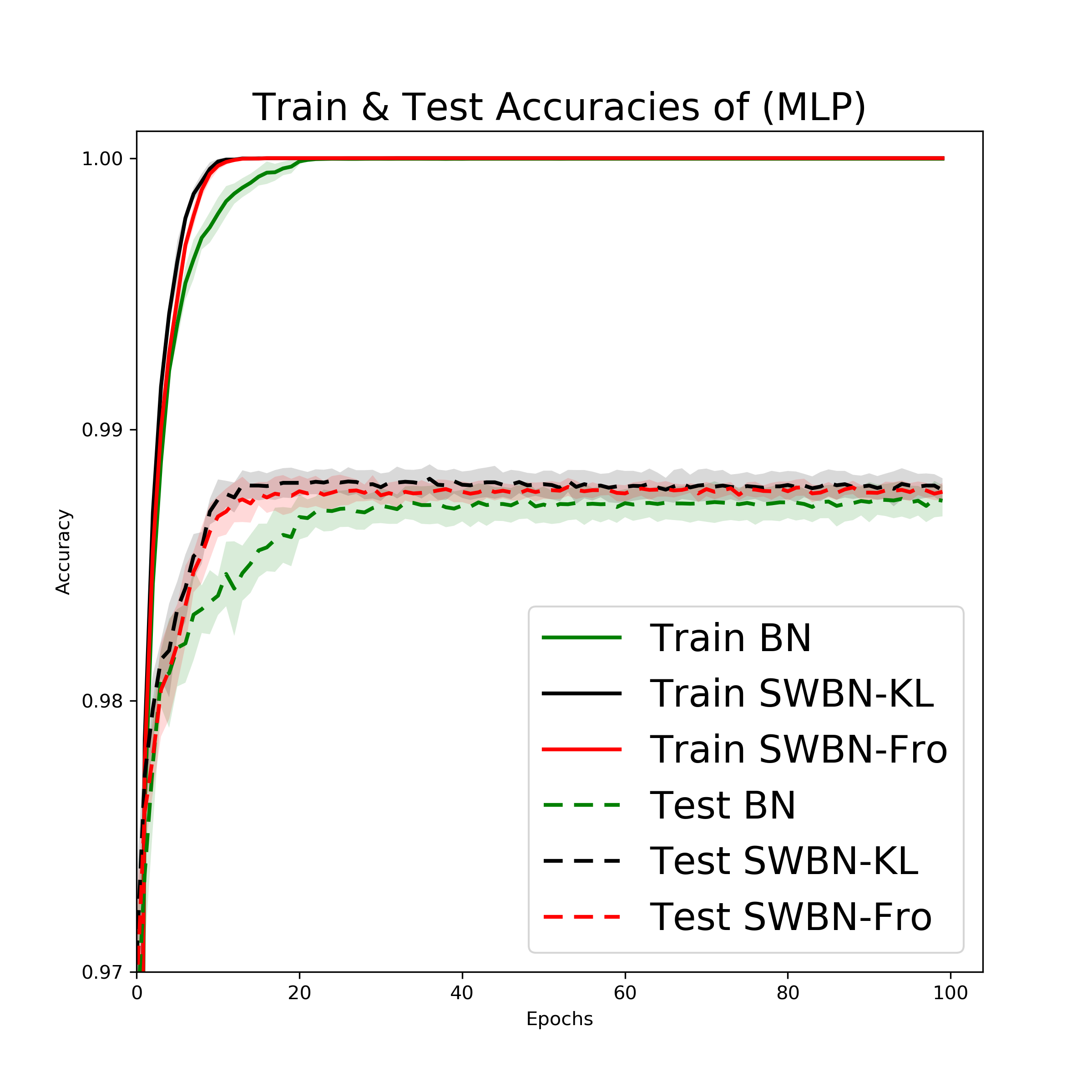}  
}
\subfigure{
\includegraphics[width=0.45\linewidth]{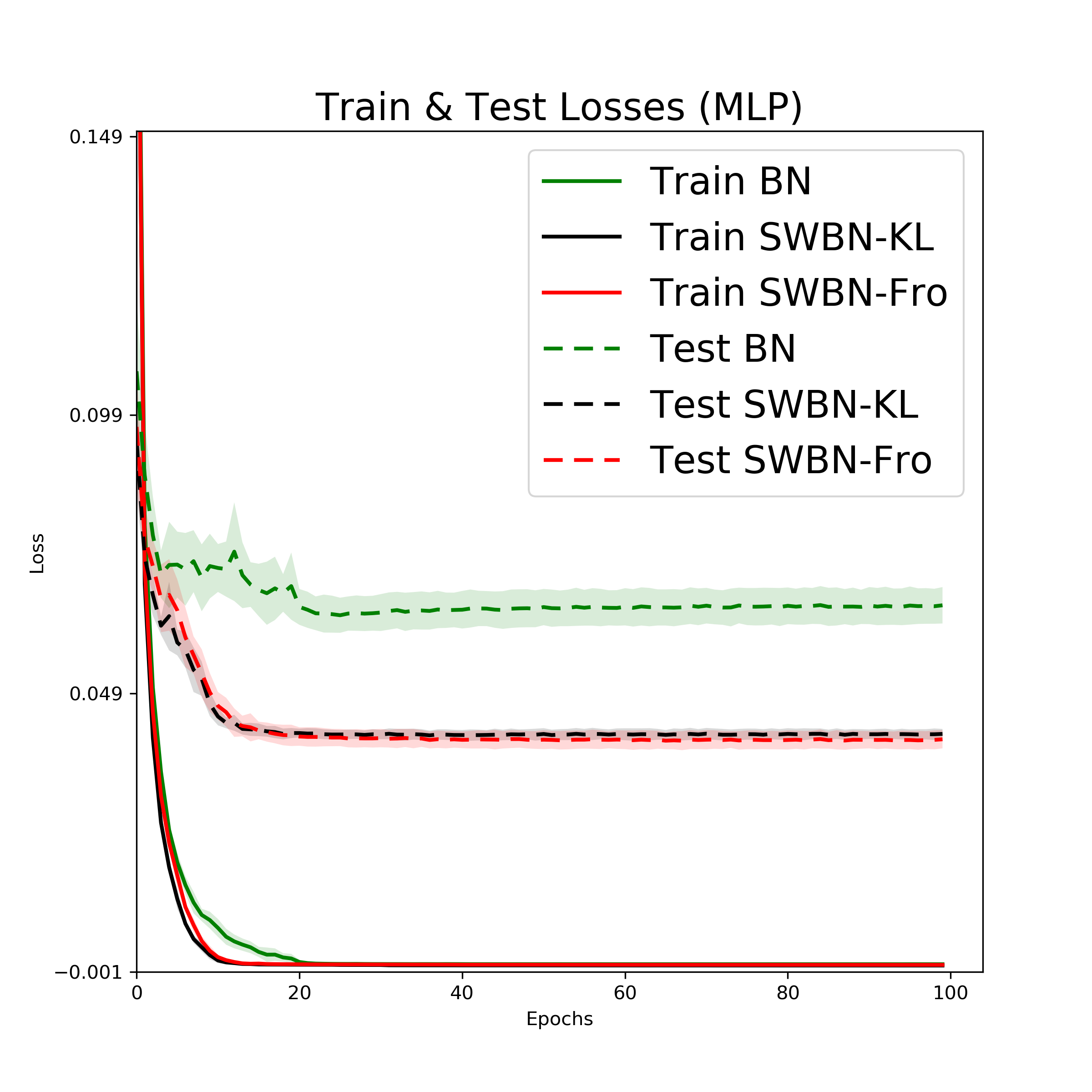}    
}
\caption{Results of Fully-connected Neural Networks on MNIST. The plots show the mean of 10 runs with $\pm1$ standard deviation. Best viewed in color.}
\label{fig:mlp_comparison}
\end{figure}

\begin{figure}
  \centering
  \subfigure{
    \includegraphics[width=0.3\linewidth]{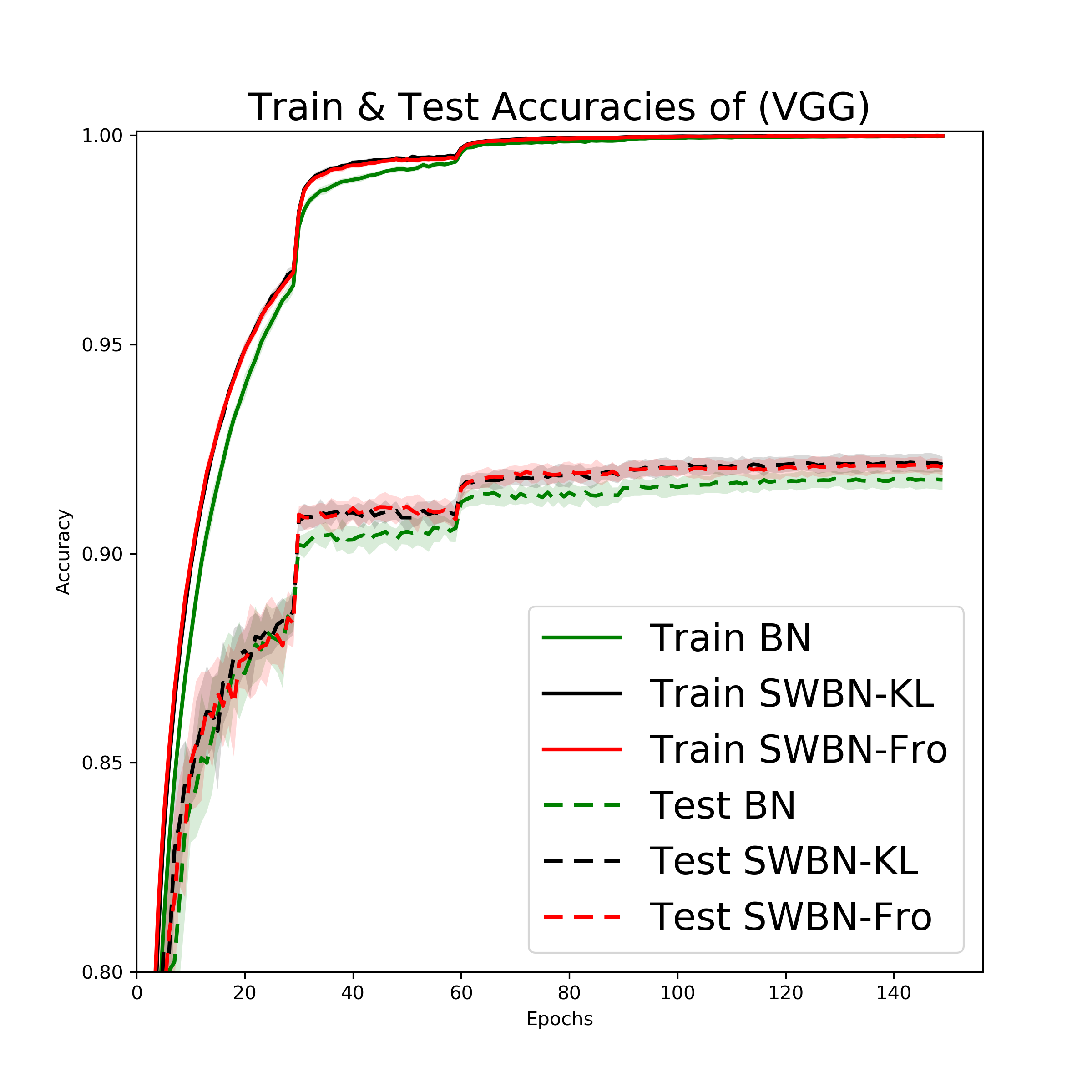}
  }
  \subfigure{
    \includegraphics[width=0.3\linewidth]{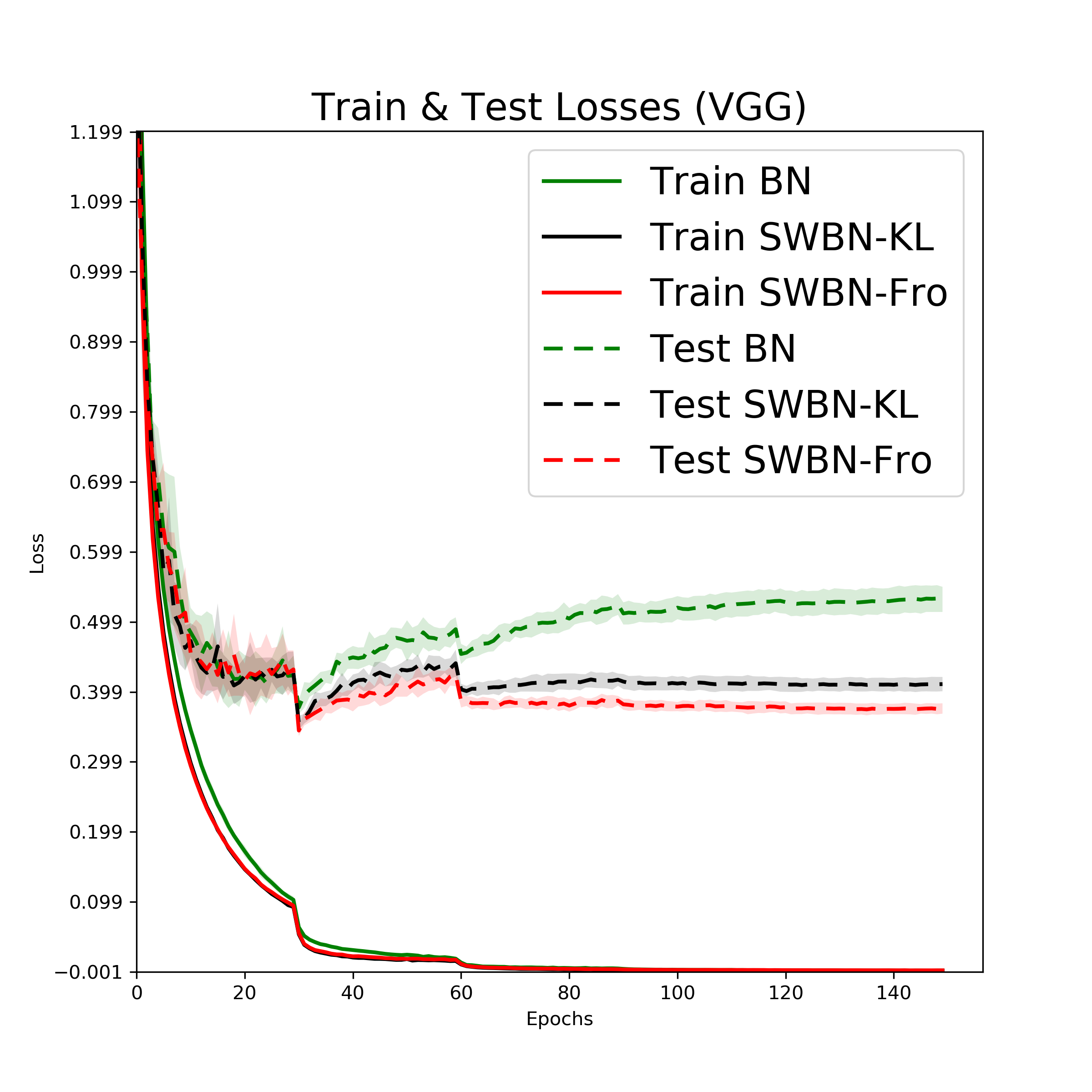}
  }\\
  \subfigure{
    \includegraphics[width=0.3\linewidth]{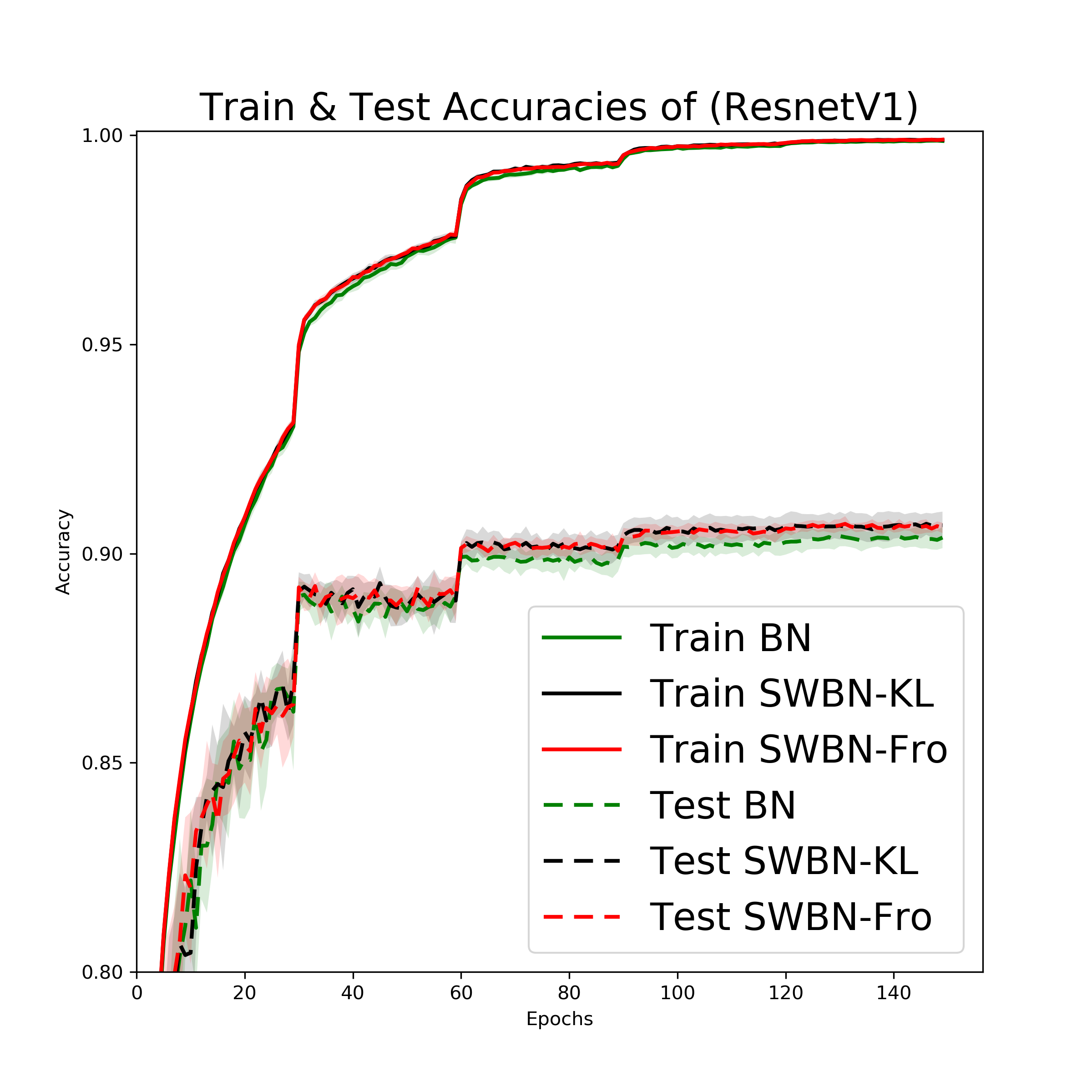}
  }
  \subfigure{
    \includegraphics[width=0.3\linewidth]{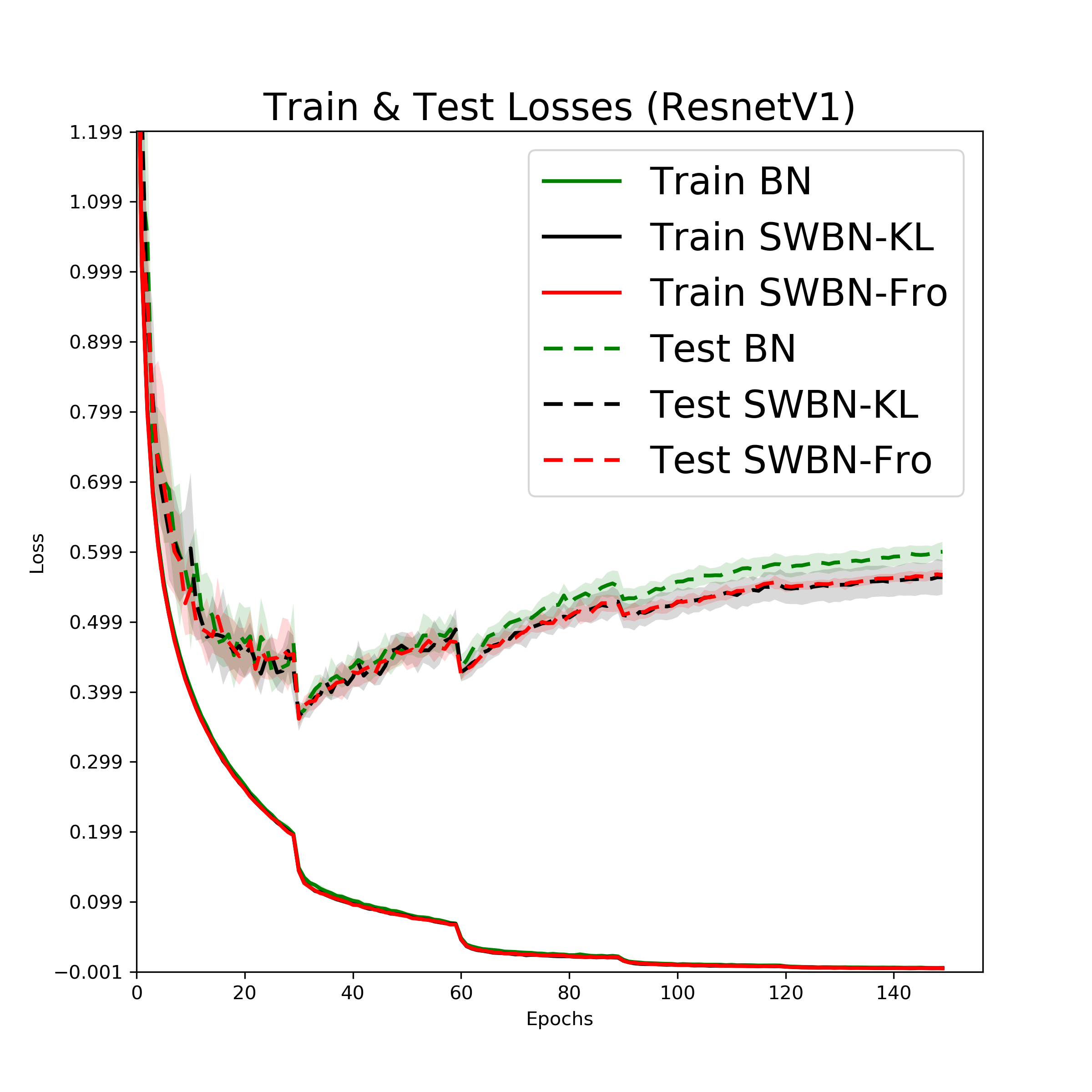}
  }
  \caption{(a,b) Results of VGG, and (c,d) ResNetV1 on CIFAR-$10$ dataset. The plots show the mean of 10 runs with $\pm1$ standard deviation. Best viewed in color.}
  \label{fig:cnn_comparison}
\end{figure}
Figures ~\ref{fig:mlp_comparison} and~\ref{fig:cnn_comparison} show the loss and accuracy plots of $10$ runs, each with a different random seed, for MNIST and CIFAR-$10$ datasets, respectively. The mean curves with $\pm1$ standard deviation are plotted for both training and test phases. As plotted, models with \ac{SWBN} layers consistently converge faster and reach lower test loss and higher accuracies than the ones with \ac{BN} layers. The results clearly indicate the superiority of \ac{SWBN} over \ac{BN}.